\documentclass[letterpaper]{article} 
\usepackage{aaai2026}  
\usepackage{times}  
\usepackage{helvet}  
\usepackage{courier}  
\usepackage[hyphens]{url}  
\usepackage{graphicx} 
\usepackage{natbib}  
\usepackage{caption} 
\usepackage{algorithm}
\usepackage{algorithmic}

\usepackage{newfloat}
\usepackage{listings}
\DeclareCaptionStyle{ruled}{labelfont=normalfont,labelsep=colon,strut=off} 
\floatstyle{ruled}
\newfloat{listing}{tb}{lst}{}
\floatname{listing}{Listing}
\usepackage{float}
\usepackage{siunitx}
\usepackage{amssymb}
\usepackage{amsmath}
\usepackage{graphicx}
\usepackage{booktabs}
\usepackage{makecell}
\usepackage{subfigure} 
\usepackage{array} 
\usepackage{multirow}
\usepackage{bbding}
\usepackage{bm}
\usepackage{amssymb}
\usepackage{pifont}
\usepackage{url}
\usepackage{amsthm}
\usepackage{tabularx}
\newtheorem{definition}{Definition}

\title{UniHR: Hierarchical Representation Learning for Unified Knowledge Graph\\ Link Prediction}

\author{
Zhiqiang Liu\textsuperscript{\rm1,3}, Yin Hua\textsuperscript{\rm1,3}, Mingyang Chen\textsuperscript{\rm4}, Yichi Zhang\textsuperscript{\rm1,3}, Zhuo Chen\textsuperscript{\rm1,3},\\ Lei Liang\textsuperscript{\rm2,3}, Wen Zhang\textsuperscript{\rm1,3}\thanks{Corresponding author.} 
}
\affiliations{\textsuperscript{\rm1}Zhejiang University\\ \textsuperscript{\rm2}Ant Group\\ \textsuperscript{\rm3}ZJU-Ant Group Joint Lab of Knowledge Graph\\
\textsuperscript{\rm4}Baichuan Inc
\\
{\{zhiqiangliu,zhang.wen\}@zju.edu.cn}}

\usepackage{bibentry}

\begin{document}

\maketitle

\begin{abstract}
Real-world knowledge graphs (KGs) contain not only standard triple-based facts, but also more complex, heterogeneous types of facts, such as hyper-relational facts with auxiliary key-value pairs, temporal facts with additional timestamps, and nested facts that imply relationships between facts. These richer forms of representation have attracted significant attention due to their enhanced expressiveness and capacity to model complex semantics in real-world scenarios. However, most existing studies suffer from two main limitations: (1) they typically focus on modeling only specific types of facts, thus making it difficult to generalize to real-world scenarios with multiple fact types; and (2) they struggle to achieve generalizable hierarchical (inter-fact and intra-fact) modeling due to the complexity of these representations. To overcome these limitations, we propose \textbf{UniHR}, a \textbf{Uni}fied \textbf{H}ierarchical \textbf{R}epresentation learning framework, which consists of a learning-optimized Hierarchical Data Representation (HiDR) module and a unified Hierarchical Structure Learning (HiSL) module. The HiDR module unifies hyper-relational KGs, temporal KGs, and nested factual KGs into triple-based representations. Then HiSL incorporates intra-fact and inter-fact message passing, focusing on enhancing both semantic information within individual facts and enriching the structural information between facts. To go beyond the unified method itself, we further explore the potential of unified representation in complex real-world scenarios. Extensive experiments on 9 datasets across 5 types of KGs demonstrate the effectiveness of UniHR and highlight the strong potential of unified representations. 
\end{abstract}
\begin{links}
\link{Code}{https://github.com/zjukg/UniHR}
\end{links}

%

\begin{figure}
\centering
\includegraphics[scale=0.47]{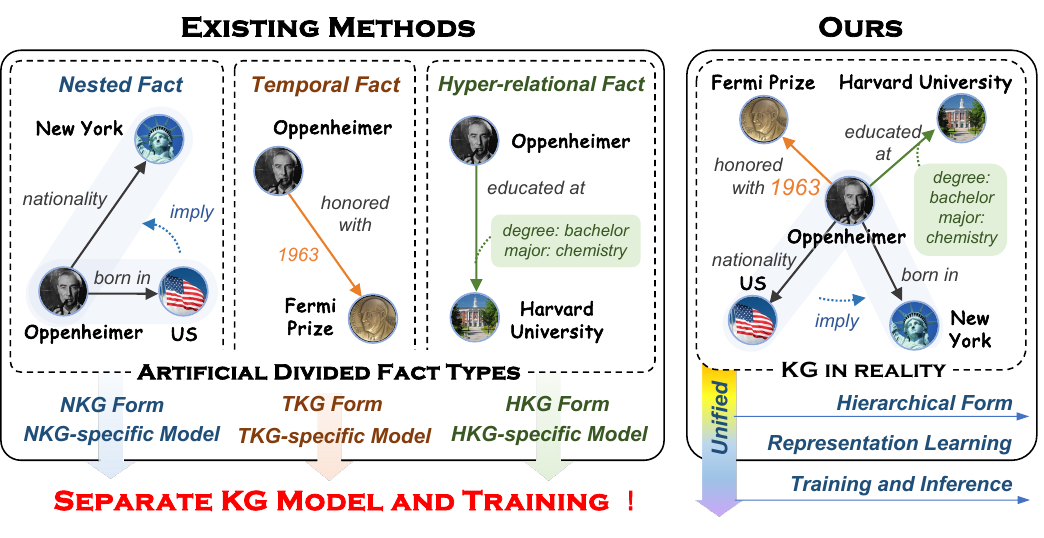}
\vspace{-2mm}
\caption{Comparison between the existing link prediction methods for specific beyond-triple KGs and our unified methods for more realistic KGs with multiple fact types.} \label{int1}
\vspace{-3mm}
\end{figure}

\section{Introduction}
Real-world large-scale knowledge graphs (KGs) such as Wikidata~\citep{wikidata} and DBpedia~\citep{DBpedia} have been widely applied in many areas including question answering~\citep{kaiser2021reinforcement} and natural language processing~\citep{annervaz2018learning,ontotune}. To faithfully represent complex real-world knowledge, these KGs usually incorporate not only standard triple-based facts, but also more complex and heterogeneous types of facts such as hyper-relational, temporal and nested facts. 

Despite the simplicity of triple-based representation (i.e., $(head,relation,tail)$), such forms struggle to capture the complexity of real-world facts, e.g., ``\textit{Oppenheimer is educated at Harvard University for a bachelor degree in chemistry}". Consequently, recent studies~\citep{beyond_triple} have focused on semantically richer beyond-triple facts, including: hyper-relational fact ((\textit{Oppenheimer, educated at, Harvard University})\textit{, degree: bachelor, major: chemistry}), temporal fact (\textit{Oppenheimer, honored with, Fermi Prize, 1963}), nested fact ((\textit{Oppenheimer, born in, New York})\textit{, imply, }(\textit{Oppenheimer, nationality, The United States})). These fact types allow for expression of complex semantics and revelation of relationships between facts. Thus in recent years, Hyper-relational KGs (HKG)~\citep{hahe}, Temporal KGs (TKG)~\citep{5EL}, and Nested factual KGs (NKG)~\citep{GRADATE} attract wide research interests. 

As shown in Figure~\ref{int1}, we find that existing research mainly suffers from two major limitations: (1) They \textbf{fail to reflect real-world scenarios with multiple heterogeneous fact types~\citep{beyond_triple}, instead artificially dividing and only modeling a single KG type}; (2) Earlier triple-based studies~\cite{hitter} have demonstrated the effectiveness of hierarchical fact semantic modeling (inter-fact and intra-fact). But due to the complexity of beyond-triple representations, they \textbf{struggle to achieve comprehensive hierarchical semantic modeling, even generalizing to other fact types}. Specifically, for HKGs, StarE~\citep{stare} customizes GNN to enhance inter-fact interactions, while GRAN~\citep{gran} designs attention variants with edge-bias to capture intra-fact heterogeneity. For NKGs, NestE~\citep{neste} et al. NKG methods connect bi-level facts and only score intra-fact semantics. For TKGs, either by explicitly incorporating temporal information into the score function like GeomE+~\citep{TGeomE+}, or by unfolding entity neighborhood subgraphs along temporal chain and capturing inter-fact semantics to model temporal information like ECEformer~\citep{eceformer}. Although advanced methods like HAHE~\citep{hahe} begin to capture hierarchical semantics for HKGs, their heterogeneity in representation limits their scalability to other fact types. \textbf{Therefore, establishing a unified hierarchical representation learning method for real-world KG with multiple fact types is worth investigating.}

To fill this research gap, we propose \textbf{UniHR}, a \textbf{Uni}fied \textbf{H}ierarchical \textbf{R}epresentation learning framework, which includes a \underline{Hi}erarchical \underline{D}ata \underline{R}epresentation (HiDR) module and a \underline{Hi}erarchical \underline{S}tructure \underline{L}earning (HiSL) module.
HiDR module standardizes hyper-relational facts, nested factual facts, and temporal facts into the form of triples without loss of information. 
Furthermore, HiSL module captures local semantic information during intra-fact message passing and then utilizes inter-fact message passing to enrich the global structure information to obtain better node embeddings based on HiDR form. Finally, the updated embeddings are fed into decoders for link prediction. Apart from the unification of method itself, UniHR's unified representation enables flexible extensions. Unlike previous KG-specific models, UniHR accommodates more complex scenarios, such as compositional knowledge graphs, multi-task learning, and joint training on hybrid facts, thereby paving the way for pre-trained models across diverse KG types. 
Our contributions can be summarized as follows.
\begin{itemize}
\item We emphasize the value of investigating unified KG representation learning method, including unified symbolic representation and unfied learning for different KGs. 
\item To our knowledge, we propose the first unified KG representation learning framework UniHR, across different types of KGs, including a hierarchical data representation module and a hierarchical  structure learning module.
\item We conduct link prediction experiments on 9 datasets across 5 types of KGs. Compared to KG-specific methods, UniHR achieves the best or competitive results, verifying strong generalization capability.
\end{itemize}

\section{Preliminaries}

\paragraph*{Link Prediction on Triple-based KG.}
A triple-based KG $\mathcal{G}_{KG}\,\text{=}\,\{ \mathcal{V} ,\mathcal{R} ,\mathcal{F} \}$ represents facts as triples, denoted as $\mathcal{F}\,\text{=}\{\left( h,r,t \right) |h,t\in \mathcal{V},$
$ r\in \mathcal{R}\} $, where $\mathcal{V}$ is the set of entities and $\mathcal{R}$ is the set of relations. The link prediction on triple-based KGs involves answering a query $\left( h,r,? \right) $ or $\left( ?,r,t \right) $, where the missing element `$?$' is an entity in $\mathcal{V}$.

\paragraph*{Link Prediction on Hyper-relational KG.}
A hyper-relational KG (HKG) $\mathcal{G}_{\text{HKG}}\,\text{=}\,\{\mathcal{V} ,\mathcal{R} ,\mathcal{F} \}$ consists of hyper-relational facts (H-Facts) $\mathcal{F}$, denoted as $\mathcal{F}\,\text{=}\{((h,r,t),$
$\{(k_i\text{:}\,v_i)\} _{i=1}^{m})|\,h,t,v_i\in\mathcal{V},r,k_i\in\mathcal{R} \}$. Typically, we refer to $\left( h,r,t \right)$ as the main triple and $\left\{ \left( k_i\text{:}\,v_i \right) \right\} _{i=1}^{m}$ as $m$ auxiliary key-value pairs. Link prediction on HKGs aims to predict entities in the main triple or the key-value pairs. Symbolically, the aim is to predict the missing element, denoted as `?' for queries $((h,r,t),(k_1\text{:}\,v_1),\ldots(k_i\text{:}\,?)),$
$((?,r,t),\{(k_i\text{:}v_i)\}_{i=1}^{m})$ or $((h,r,?),\{(k_i\text{:}v_i)\}_{i=1}^{m})$.

\paragraph*{Link Prediction on Nested Factual KG.}
A nested factual KG (NKG) can be represented as $\mathcal{G}_{\text{NKG}}\,\text{=}\,\{ \mathcal{V} ,\mathcal{R} ,\mathcal{F} ,\hat{\mathcal{R}},\hat{\mathcal{F}}\}$, which is composed of two levels of facts, called atomic facts and nested facts. $\mathcal{F}\, \text{=}\left\{ \left( h,r,t \right) |h,t\in \mathcal{V} ,r\in \mathcal{R} \right\} $ is the set of atomic facts, where $\mathcal{V}$ is a set of atomic entities and $\mathcal{R}$ is a set of atomic relations. $\hat{\mathcal{F}}\,\text{=}\,\{ \left( \mathcal{F} _i,\hat{r},\mathcal{F} _j \right)|$
$\mathcal{F} _i,\mathcal{F} _j\in \mathcal{F} ,\hat{r}\in \hat{\mathcal{R}}\} $ is the set of nested facts, where $\hat{\mathcal{R}}$ denotes nested relations.
We refer to the link prediction on atomic facts as \textit{Base Link Prediction}, and the link prediction on nested facts as \textit{Triple Prediction}. For base link prediction, given a query $\left( h,r,? \right) $ or $\left( ?,r,t \right) $, the aim is to predict missing atomic entity `$?$' from $\mathcal{V}$. For triple prediction, given a query $\left( ?,\hat{r},\mathcal{F} _j \right) $ or $\left( \mathcal{F} _i,\hat{r}, ? \right)$, the aim is to predict atomic fact `$?$' from $\mathcal{F}$.

\paragraph{\textbf{Link Prediction on Temporal KG.}}
A temporal KG (TKG) $\mathcal{G}_{\text{TKG}}\,\text{=}\,\{ \mathcal{V} ,\mathcal{R},\mathcal{F},\mathcal{T}\}$ is composed of quadruple-based facts, represented as $\mathcal{F}\,\text{=}\,\{(h,r,t,[\tau_b,\tau_e] )|h,t \in \mathcal{V},$
$r\in \mathcal{R} ,\tau_b,\tau_e\in \mathcal{T}\} $, where $\tau_b$ is the begin time, $\tau_e$ is the end time, $\mathcal{V}$ is the set of entities, $\mathcal{R}$ is the set of relations and $\mathcal{T}$ is the set of timestamps. The link prediction on TKGs aims to predict missing entities `$?$' in $\mathcal{V}$ for two types of queries $\left( ?,r,t,[\tau_b,\tau_e] \right)$ or $\left( h,r,?,[\tau_b,\tau_e]  \right)$.


\section{Related Works}

\begin{figure*}
\centering
\includegraphics[scale=0.61]{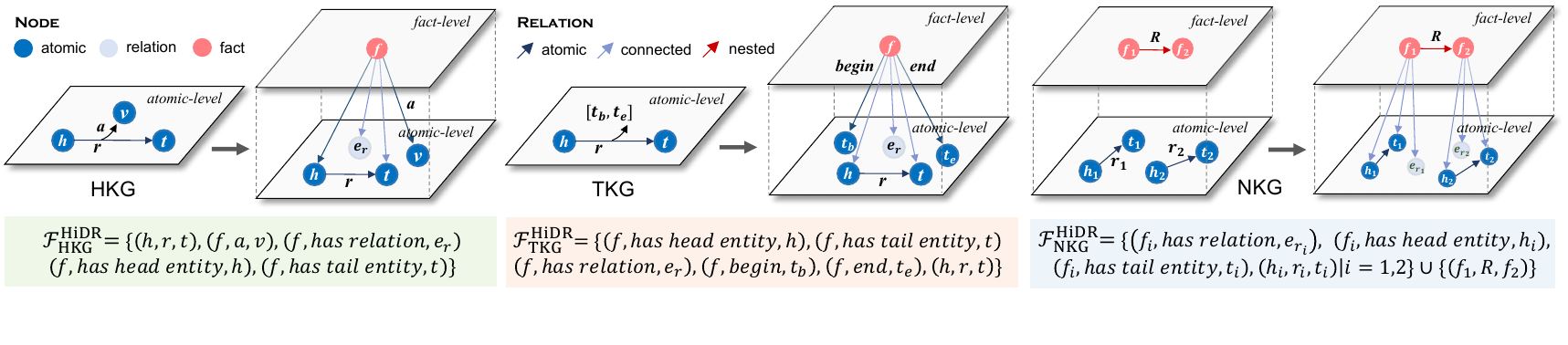}
\vspace{-11mm}
\caption{Diverse beyond-triple facts are translated into the hierarchical data representation (HiDR) form.}
\vspace{-3mm}
\label{fig1}
\end{figure*}

\paragraph{Link Prediction on Hyper-relational Knowledge Graph.}
Early HKG methods typically focus on modeling either local or global information.~\citet{stare} customize StarE based on CompGCN~\citep{CompGCN} for H-Facts to capture global structure information among H-Facts, demonstrating the importance of structure information in HKGs. GRAN~\citep{gran} with edge-aware bias in attention~\citep{transformer} layer, HyNT~\citep{hynt} with qualifier-aware encoder, and ShrinkE with relation-specific box~\citep{shrinke} all focus on modeling intra-fact semantic information. Recent advanced methods aim to comprehensively capture both inter-fact and intra-fact information. For example, HAHE~\citep{hahe} employs dual-graph attention and edge-aware bias in the transformer attention layer for hierarchical modeling, achieving significant performance improvements. Similarly, HyperSAT~\citep{HyperSAT} accomplishes this through a combination of graph sampling and a key-value joint attention mechanism. We consider hierarchical modeling of KGs to be a promising direction, existing approaches are customized for HKG form and difficult to generalize to other types of facts.

\paragraph{Link Prediction on Nested Factual Knowledge Graph.}
Chung et al.~\citep{bive} are the first to introduce nested facts to model relationships between facts, and also propose BiVE which bridges semantics between atomic facts and fact nodes via a simple MLP and scores both atomic facts and nested facts using quaternion-based KGE scoring functions like QuatE~\citep{quate} or BiQUE~\citep{bique}. Based on BiVE, NestE~\citep{neste} represents fact nodes using a $1 \times 3$ embedding matrix and the nested relations as a $3\times 3$ matrix to avoid information loss. GRADATE~\citep{GRADATE} enhances entity and fact representation learning by mining latent intra-fact semantics. However, due to the complexity of NKG representations, existing methods have so far primarily focused on capturing intra-fact semantic information.

\paragraph{Link Prediction on Temporal Knowledge Graph.}Recent studies on TKG representation learning have mainly focused on designing elegant time-aware modules to enhance representation capability. Advanced models like TGeomE+~\citep{TGeomE+} improve the modeling of local semantics in TKGs through 4th-order tensor factorization and linear temporal regularization. Similarly, HGE~\citep{HGE} and 5EL~\citep{5EL} enhance the expressiveness of temporal spaces by introducing geometric spaces. In contrast, ECEformer~\citep{eceformer} captures only inter-fact semantics by unfolding entity neighborhood subgraphs along the temporal chain to model temporal information. However, TKG representation learning methods that can simultaneously capture both intra-fact temporal semantics and global structural information remain largely unexplored. UniHR achieves this by regarding timestamps as nodes and directly employing hierarchical message passing.


\section{Methodology}
In this section, we introduce \textbf{UniHR}, a \textbf{Uni}fied \textbf{H}ierarchical \textbf{R}epresentation learning framework, which includes a \underline{Hi}erarchical \underline{D}ata \underline{R}epresentation (HiDR) module and a \underline{Hi}erarchical \underline{S}tructure \underline{L}earning (HiSL) module. Our workflow includes the following three steps: 
\textbf{(1)} Given a KG $\mathcal{G}$ of any type, we represent it into HiDR form $\mathcal{G}^{\text{HiDR}}$. \textbf{(2)} The $\mathcal{G}^{\text{HiDR}}$ will be encoded by HiSL module to enhance the semantic information within individual facts and structural information between facts on the whole graph. \textbf{(3)} In the phase of decoding, the updated node and edge embeddings are serialized and fed into the transformer to optimize the model.

\subsection{Hierarchical Data Representation}
To overcome the limitations of beyond-triple representations in hierarchical modeling, we introduce a \underline{Hi}erarchical \underline{D}ata \underline{R}epresentation module (\textbf{HiDR}), which is optimized for representation learning. 

Unlike existing triple-based systems~\citep{RDF} including RDF (triple) reification, RDF-star and labeled RDF representation techniques, we constrain ``triple" to serve as the basic units of HiDR form, and split ``nodes'' and ``relations'' into three types respectively, making it more suitable for graph representation learning. Meanwhile, ``triple" form makes HiDR could continuously benefit from the model developments of triple-based KGs, which is the most active area of research for link prediction over KGs. 

As shown in Figure~\ref{fig1}, we denote the entities in original KGs as \textit{atomic nodes} and abstract \textit{fact nodes} for HKGs and TKGs lacking a designated fact node. To facilitate the interaction between fact nodes and relations explicitly, we incorporate \textit{relation nodes} into the graph, represented as $e_r$ for each relation $r$. To facilitate direct access of fact nodes to the relevant atomic nodes during message passing, we also introduce three \textbf{\textit{connected relations}}: \textit{has relation}, \textit{has head entity} and \textit{has tail entity}, which establish directly connections between atomic nodes and fact nodes. Ultimately, we denote the (main) triple $\left( h,r,t \right) $ in original fact as three \textit{connected facts}: $\left( f,has\,relation,e_r \right), \left( f,has\,head\,entity,h \right),(f,has\,tail\,$

\noindent$entity,t)$, and an \textit{atomic fact} $(h,r,t)$, where $f$ is \textit{fact node}. Formally, the definition of HiDR form is as follows:

\begin{definition}
Hierarchical Data Representation (HiDR): A KG represented as the HiDR form is denoted as $\mathcal{G}^{\text{HiDR}}\,\text{=}\,$

\noindent$\{\mathcal{V}^{\text{HiDR}}, \mathcal{R}^{\text{HiDR}}, \mathcal{F}^{\text{HiDR}}\},$ where $\mathcal{V}^{\text{HiDR}}\,\text{=}\,\mathcal{V}_a\cup\mathcal{V}_r\cup\mathcal{V}_f$ is a joint set of atomic node set ($\mathcal{V}_a$), relation node set ($\mathcal{V}_r$), fact node set ($\mathcal{V}_f$). $\mathcal{R}^{\text{HiDR}}\,\text{=}\,\mathcal{R}_a \cup \mathcal{R}_n \cup \mathcal{R}_c$ is a joint set of atomic relation set ($\mathcal{R}_a$), nested relation set ($\mathcal{R}_n$), connected relation set $\mathcal{R}_c\,\text{=}\{has\,relation, has\,head\,entity,has\,tail\,entity\}$. The fact set $\mathcal{F}^{\text{HiDR}}\,\text{=}\,\mathcal{F}_a \cup \mathcal{F}_c \cup \mathcal{F}_n$ is jointly composed of three types of triple-based facts: atomic facts ($\mathcal{F}_a$), connected facts ($\mathcal{F}_c$) and nested facts ($\mathcal{F}_n$), where $\mathcal{F}_a\text{=}\,\{(v_1,r,$

\noindent$v_2) |\,v_1, v_2 \in \mathcal{V}_a, r \in \mathcal{R}_a\}$, $\mathcal{F}_c\,\text{=}\,\{ (v_1, r, v_2) |\,v_1 \in \mathcal{V}_f,r \in \mathcal{R}_c, v_2 \in \mathcal{V}_a\}$, $\mathcal{F}_n\,\text{=}\,\{ (v_1, r, v_2) |\,v_1, v_2 \in \mathcal{V}_f, r \in \mathcal{R}_n\}$. 
\end{definition}

\begin{figure*}
\centering
\includegraphics[scale=0.61]{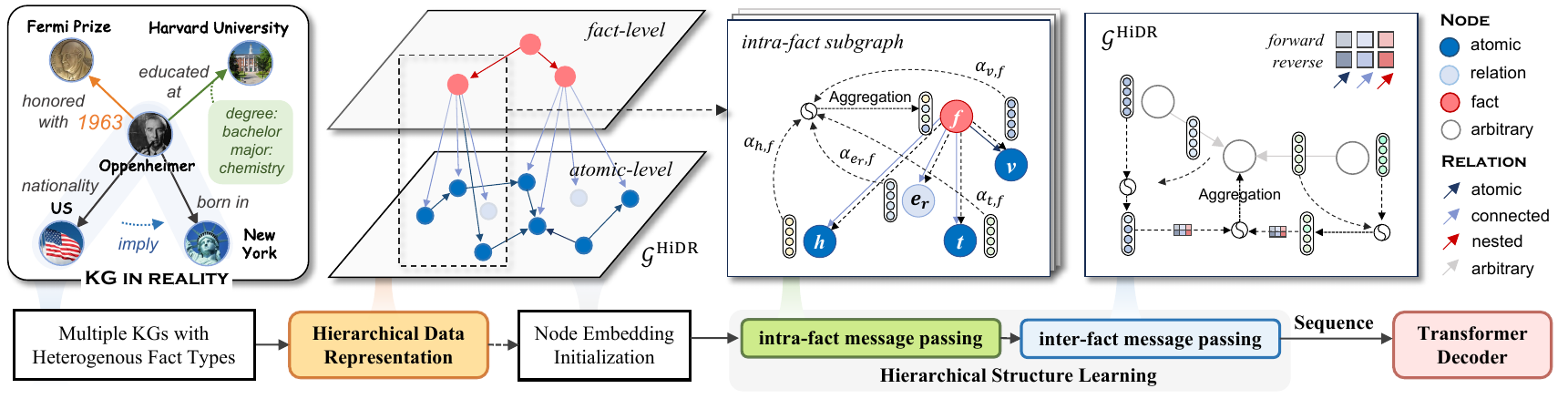}
\vspace{-6mm}
\caption{The overview of UniHR, including HiSL module for intra-fact and inter-fact message passing.} \label{fig2}
\vspace{-3mm}
\end{figure*}

Next, we introduce how to transform different types of beyond-triple KGs into HiDR form. 

\noindent\textbf{For HKGs,} we regard the key-value pairs as complementary information for facts. Thus, we translate the H-Facts $\mathcal{F}_{\text{HKG}}\,\text{=}\{((h,r,t),\{( k_i\text{:}\,v_i)\} _{i=1}^{m}\}$ into the HiDR form that $\mathcal{G}^{\text{HiDR}}_{\text{HKG}}\,\text{=}\,\{ \mathcal{V},\mathcal{R}, \mathcal{F}^{\text{HiDR}}_{\text{HKG}}\}$ following the definition, where $\mathcal{F}_{c}\,\text{=}$

\noindent$\{(f,has\,relation,e_r),(f, has\,head\,entity,h),(f,has\,tail\,$

\noindent$entity,t),(f, k_1, v_1),\ldots, (f, k_m, v_m)\}$, $\mathcal{F}_a\text{=}\{ (h,r,t)\,|((h,$

\noindent$r,t),\{(k_i\text{:}\,v_i)\}_{i=1}^{m}) \in \mathcal{F}_{\text{HKG}})\}$ and $\mathcal{F}_n\text{=}\,\varnothing $. 

\noindent\textbf{For NKGs,} HiDR can naturally represent the hierarchical facts, so we translate the atomic facts $\mathcal{F}_{\text{NKG}}\,\text{=}\,\{(h_i,r_i,t_i)\}$ and the nested facts $\hat{\mathcal{F}}_{\text{NKG}}\,\text{=}\,\{(( h_1,r_1,t_1),R,(h_2,r_2,t_2))|$

\noindent$(h_i,r_i,t_i)\in\mathcal{F}_{\text{NKG}}\}$ into the form of HiDR that $\mathcal{G}^{\text{HiDR}}_{\text{NKG}}\,\text{=}\,\{\mathcal{V},$

\noindent$\mathcal{R},\mathcal{F}^{\text{HiDR}}_{\text{NKG}}\}$ following the definition, where $\mathcal{F}_a\,\text{=}\,\{( h_i,r_i,t_i)|$

\noindent$( h_i,r_i,t_i)\in\mathcal{F}_{\text{NKG}}\}$, $\mathcal{F}_c\,\text{=}\{( f_i,has\,head\,entity,h_i),(f_i,$

\noindent$has\,tail\,entity,t_i),( f_i,has\,relation,e_{r_i} )|f_i\,\text{=}\,( h_i,r_i,t_i)\in$
$\mathcal{F}_{\text{NKG}}\}$ and $\mathcal{F}_n\text{=}\{(f_1,R,f_2)|f_i\in\mathcal{F}_{\text{NKG}}\}$. 

\noindent\textbf{For TKGs,} we regard the TKG as a special HKG, and convert timestamps to auxiliary key-value pairs in HKGs by adding two special \textit{atomic relations}: \textit{begin} and \textit{end}, regarding timestamps as special numerical atomic nodes. Thus, we firstly translate the temporal facts in TKGs $\mathcal{F}_{\text{TKG}}\text{=}\,\{(h,r,t,[\tau_b,\tau_e])\}$ into H-Facts form $\mathcal{F}_{\text{TKG}}^{\text{HKG}}\text{=}\,\{( h,r,$

\noindent$t,begin\text{:}\tau_b,end\text{:}\tau_e)\}$. Then according to the previous transformation in HKG, it can be translated into the HiDR form that $\mathcal{G}_{\text{TKG}}^{\text{HiDR}}\text{=}\{\mathcal{V},\mathcal{R},\mathcal{F}_{\text{TKG}}^{\text{HiDR}}\}$ following the definition, where $\mathcal{F}_a\,\text{=}\,\{(h,r,t)\,|\,( h,r,t,begin\text{:}\,\tau_b,end\text{:}\tau_e)\in\mathcal{F}_{\text{TKG}}^{\text{HKG}}\}$, $\mathcal{F}_c\,\text{=}$

\noindent$\{(f,has\,relation,e_r),(f,has\,head\,entity,h),\,( f,has\,tail\,$

\noindent$entity,t),\,(f,begin,\tau_b),( f,end,\tau_e)\,|\,f\,\text{=}\,( h,r,t,begin\text{:}\,\tau_b,$

\noindent$end\text{:}\,\tau_e )\in\mathcal{F}_{\text{TKG}}^{\text{HKG}}\}$ and $\mathcal{F}_n\, \text{=}\, \varnothing $.

In summary, HiDR serves as a module that dynamically transforms the original data into a unified representation optimized for the model, without altering the storage format of the original data. Moreover, from the perspective of graph learning, it fully preserves the semantics of the original knowledge graphs without any loss of information.

\subsection{Hierarchical Structure Learning}
It's evident that HiDR form introduces many additional relation nodes and fact nodes. To avoid significantly increasing the model's training parameters while fully capturing the hierarchy of HiDR form, we design a \underline{Hi}erarchical \underline{S}tructure \underline{L}earning module, abbreviated as \textbf{HiSL} shown in Figure~\ref{fig2}.

\paragraph{\textbf{Representation Initialization.}}
We ﬁrst initialize the embedding matrices $\mathbf{H}_{a}\in \mathbb{R} ^{|\mathcal{V}_a|\times d}$ and $\mathbf{E}\in \mathbb{R} ^{|\mathcal{R}^{\text{HiDR}}|\times d}$ for atomic nodes and all relation edges. Then we also initialize the embedding of relation node $\mathbf{H}_{r}\in \mathbb{R} ^{|\mathcal{V}_r|\times d}$, which can be transformed from the relation edge $r$ with a projection matrix $\mathbf{W_r}\in \mathbb{R} ^{d\times d}$: $\mathbf{H}_{r}=\mathbf{E_a\cdot W_r}$, where $\mathbf{E_a}\subseteq \mathbf{E}$ is the atomic relation embeddings. Then we initialize the fact node embeddings $\mathbf{H}_{f}$ to explicitly capture key information within facts by utilizing the embedding of (main) triple: 
\begin{equation}
    \mathbf{h}_{f}=f_{m}([\mathbf{h}_{h};\mathbf{h}_{r};\mathbf{h}_{t}]),
\end{equation}
\noindent where $(h,r,t)\in \mathcal{F}_{a}$, $f_{m}\text{:}\;\mathbb{R}^{3d}\rightarrow \mathbb{R}^{d}$ is a 1-layer MLP, $[\cdot; \cdot]$ is the concatenation, $\mathbf{h}_{h},\mathbf{h}_{t}\subseteq \mathbf{H}_{a},\mathbf{h}_{r}\subseteq \mathbf{H}_{r}$ denote (main) triple embedding. Therefore, the initialization of relation nodes and fact nodes is sufficiently parameter-efficient.

For numerical atomic nodes, namely timestamps in temporal knowledge graphs, we encode the timestamp $\tau$ into an embedding with Time2Vec~\citep{Time2Vec}: $\mathbf{h}_{\tau}=\omega_{p}\sin \left(f_p(\tau)\right)+f_{np}(\tau)$, where $f_{p}\text{:}\;\mathbb{R}^{1}\rightarrow \mathbb{R}^{d}$ and $f_{np}\text{:}$$\mathbb{R}^{1}\rightarrow \mathbb{R}^{d}$ are both 1-layer linear layers as periodic and non-periodic functions, and $\omega_{p}\in\mathbb{R}^{1}$ is a learnable parameter for scaling the periodic features.

\paragraph{\textbf{Intra-fact Message Passing.}}
In this stage, massage passing is conducted for fact nodes. Given a fact node $f_k\in\mathcal{V}_f$, we construct its constituent elements, i.e., one-hop neighbors, as the node set $\mathcal{V} _k\,\text{=}\,\{v\in \mathcal{N}_{f_k}\,|\,v\in \mathcal{V}_{a}\cup \mathcal{V}_{r}\}$, where $\mathcal{N} _{f_k}$ is the set of one-hop neighbors of fact node $f_k$. Then we retain the edges directly connected to fact node $f_k$, thereby constructing a subgraph $\mathcal{G} _k\,\text{=}\,\{ \mathcal{V}_k, \mathcal{R}_k,\mathcal{F}_k\}\subseteq\mathcal{G}^{\text{HiDR}}$. 
\begin{table*}[ht]
\renewcommand{\arraystretch}{1.0} 
\centering
\resizebox{0.99\textwidth}{!}{
\begin{tabular}{lcccccccccccc}
\toprule
              & \multicolumn{6}{c}{\textbf{WikiPeople}}                                                                      & \multicolumn{6}{c}{\textbf{WD50K}}                                                                           \\ \cmidrule(lr){2-7}\cmidrule(lr){8-13} 
\multicolumn{1}{l}{\textbf{Model}}        & \multicolumn{3}{c}{subject/object}               & \multicolumn{3}{c}{all entities}                   & \multicolumn{3}{c}{subject/object}               & \multicolumn{3}{c}{all entities}                 \\ \cmidrule(lr){2-4}\cmidrule(lr){5-7}\cmidrule(lr){8-10}\cmidrule(lr){11-13} 
              & MRR            & H@1          & H@10         & MRR            & H@1          & \multicolumn{1}{c}{H@10}         & MRR            & H@1          & H@10         & MRR            & H@1          & H@10         \\ \midrule
NaLP~\citep{NaLP}         & 0.356          & 0.271          & 0.499          & 0.360           & 0.275          & 0.503          & 0.230          & 0.170          & 0.347          & 0.251          & 0.187          & 0.375          \\
StarE~\citep{stare}       & 0.458          & 0.364          & \underline{0.611} & -               & -               & -            & 0.309          & 0.234          & 0.452          & -               & -               & -               \\
GRAN~\citep{gran}      & 0.477          & 0.408          & 0.596    & 0.480          & 0.411          & 0.599  & 0.329          & 0.259          & 0.465          & 0.363    & 0.294          & 0.493    \\
tNaLP~\citep{tnalp}         & 0.358          & 0.288          & 0.486          & 0.361          & 0.290          & 0.490          & 0.221         & 0.163          & 0.331          & 0.243          & 0.182          & 0.360          \\
HyNT~\citep{hynt}       & 0.479    & 0.411    & 0.601          & 0.478    & 0.409    & 0.601       & 0.328    & 0.256    & 0.468    & 0.356          & 0.285    & 0.493          \\  
ShrinkE~\citep{shrinke}       & 0.485    & 0.431    & 0.601   & -    & -    & -    & 0.345          & 0.275    & 0.482    & -    & -   & -        \\ 
HAHE$^{*}$~\citep{hahe}       & \textbf{0.498}    & {0.418}    & 0.610          & \textbf{0.497}    & \underline{0.421}    & \underline{0.614}       & 0.343    & 0.269    & \underline{0.484}    & 0.378          & \underline{0.306}    & 0.515          \\ 
NYLON$^{*}$~\citep{nylon}   & 0.385     & 0.299    & 0.527    & 0.384    & 0.300   & 0.520    & 0.326    & 0.262    & 0.446   & 0.291    & 0.226    & 0.414            \\ 
HyperSAT~\citep{HyperSAT}       & 0.493    & \textbf{0.427}    & 0.610          & \underline{0.496}    & \textbf{0.430}    & 0.613       & \underline{0.345}    & \underline{0.270}    & \textbf{0.489}    & \underline{0.380}          & \underline{0.306}    & \textbf{0.520}          \\ \midrule
\textbf{UniHR}       & \underline{0.496} & \underline{0.419} & \textbf{0.619}    & \underline{0.496} & 0.420 & \textbf{0.621}   & \textbf{0.348} & \textbf{0.278} & {0.482} & \textbf{0.382} & \textbf{0.313} & \underline{0.517} \\ \bottomrule
\end{tabular}}
\vspace{-1.5mm}
\caption{Results on HKG datasets, $^{*}$ are reproduced by us and others are taken from~\citep{HyperSAT}. }\label{tab1}
\vspace{-3mm}
\end{table*}
For this subgraph, we employ the graph attention to aggregate local information, computing the attention score $\alpha_{i,j}$ between node $i\in\mathcal{V}_k$ and its neighbor $j$. The formula for calculating $\alpha_{i,j}$ in the $l$-th layer is as follows:
\begin{equation}\resizebox{0.75\hsize}{!}{$
\alpha _{i,j}^l=\frac{\exp \left(\mathbf{W}^l\left( \sigma \left( \mathbf{W}_{in}^l\mathbf{h}_i^l+\mathbf{W}_{out}^l\mathbf{h}_j^l \right) \right) \right)}{\sum\limits_{j'\in \mathcal{N} _i}\exp \left( \mathbf{W}^l\left( \sigma \left( \mathbf{W}_{in}^l\mathbf{h}_i^l+\mathbf{W}_{out}^l\mathbf{h}_{j'}^l \right) \right) \right)}
,$}
\end{equation}
where $\mathbf{h}_i^l,\mathbf{h}_j^l\in \mathbb{R} ^d$ represent the embeddings of node $i$ and its neighbor $j$ in $l$-th layer. And there are three learnable weight matrices  $\mathbf{W}_{in}^l,\mathbf{W}_{out}^l\in \mathbb{R} ^{d\times d}$ and $\mathbf{W}^l\in \mathbb{R} ^{d}$. We choose LeakyReLU as activation function $\sigma$. Then, the updated node embeddings are obtained by aggregating the information of neighbors according to the attention scores:
\begin{equation}\resizebox{0.5\hsize}{!}{$
\mathbf{h}_i^l=\mathbf{h}_i^l+\sum\limits_{j\in \mathcal{N} _i}\alpha _{i,j}^l\cdot \mathbf{W}_{out}^l\mathbf{h}_j^l.
$}\end{equation}
\paragraph{\textbf{Inter-fact Message Passing.}}At this stage, message passing is conducted on the whole graph $\mathcal{G}^{\text{HiDR}}$. Specifically, we use a non-parametric aggregation operator $\phi\left(\cdot\right)\text{:}\mathbb{R}^d\times\mathbb{R}^d\rightarrow \mathbb{R}^d$ to obtain messages from neighbouring nodes and edges. We employ the circular-correlation operator, defined as:
\begin{equation}
\resizebox{0.8\hsize}{!}{$
\phi \left( \mathbf{h}_j,\mathbf{e}_r \right) =\mathbf{h}_j\star \mathbf{e}_r=\mathbf{F}^{-1} \left( \left( \mathbf{F} \mathbf{h}_j \right) \odot \overline{\left( \mathbf{F} \mathbf{e}_r \right)} \right),
$}\end{equation}

\noindent where $\mathbf{F}$ and $\mathbf{F}^{-1}$ denote the discrete fourier transform (DFT) matrix and its inverse matrix, and the $\odot$ is the element-wise product. In order to fully capture the graph's heterogeneity, we classify edges along two dimensions: direction $\lambda(r)\in \left\{ forward, reverse \right\}$ and type $\tau(r)\in \{connected\,\,relation,atomic\,\,relation,nested\,\,relation\} $ and adopt two relation-specific learnable parameters $\mathbf{W}_{\lambda \left( r \right)} $

\noindent$\in\mathbb{R} ^{d\times d}$ and $\omega _{\tau \left( r \right)}\in $
$\mathbb{R} ^1$ for fine-grained aggregation: 
\begin{equation}
\resizebox{0.9\hsize}{!}{$
\mathbf{h}_{i}^{l+1}= \sum\limits_{(r,j) \in \mathcal{N}(i)}{\sigma \left( \omega _{\tau \left( r \right)}^{l} \right)\mathbf{W}_{{\lambda }\left( r \right)}^{{l}}\phi \left( \mathbf{h}_{j}^{l},\mathbf{e}_{r}^{l} \right)}+\mathbf{W}_{self}^{l}\mathbf{h}_{i}^{l} ,
$}\end{equation}

\noindent where $\mathbf{W}^l_{self}\in\mathbb{R} ^{d\times d}$, $\sigma$ is a sigmoid function and $\mathcal{N} \left( i \right)$ is a set of immediate neighbors of $i$ for its outgoing edges $r$. We utilize $\phi\left(\cdot\right)$ to combine the information from edge $r$ and node $j$, and then passes it to node $i$ for update. Meanwhile, we update the relation representation as: $\mathbf{e}_{r}^{l+1}=\mathbf{W}_{rel}^{l}\mathbf{e}_{r}^{l}$.

Through Intra-fact and Inter-fact two-stage message passing, nodes can fully capture both local semantic and global structural information. Moreover, the number of training parameters does not increase with the scale of the graph, thereby effectively adapting to the HiDR form.

\subsection{Link Prediction Decoder}
Since the query varies across different settings, we use the transformer~\citep{transformer} as the decoder with \textit{mask} pattern. Specifically, we serialize the updated node and edge embeddings into a sequence of fact embeddings, mask the elements to be predicted in facts with the $\bm{\left[ M \right]}$ token as the input. Finally, we obtain the embedding of output $\bm{\left[ M \right]}$ in the last layer to measure the plausibility of the fact, denoted as $\mathbf{h}_{pre}$, and calculate the probability distribution of candidates, followed by training it using the cross-entropy loss:
\begin{equation}
\mathcal{L} =\sum_{t=0}^{\left| \mathcal{R}\right|+\left|\mathcal{V} \right|}{y_t\log P_t},
\end{equation}
where $P=\,{\rm Softmax} \left( f\left( \mathbf{h}_{pre} \right) [\mathbf{E};\mathbf{H}]^{\top} \right)\in\mathbb{R} ^{\left| \mathcal{R} \right|+\left| \mathcal{V} \right|}$ represents the confidence scores of all candidates, $f\text{:}\; \mathbb{R} ^d\rightarrow \mathbb{R} ^d$ is a 1-layer MLP, and $[\mathbf{E};\mathbf{H}]\in\mathbb{R} ^{(\left| \mathcal{R} \right|+\left| \mathcal{V} \right|)\times d}$ is the embedding matrix of all candidate edges or nodes. The $P_t$ and $y_t$ are probability and ground truth of the $t$-th candidate.


\section{Experiment}
\subsection{Experiment Settings}

\paragraph{Datasets.}
\textit{For HKGs,} we use WikiPeople~\citep{NaLP} and WD50K~\citep{gran}. \textit{For NKGs,} we select FBH, FBHE and DBHE~\citep{bive}. \textit{For TKGs,} we use wikidata12k~\citep{hyte}. To further evaluate the potential of the unified representation, we further introduce hyper-relational TKG datasets WIKI-hy and YAGO-hy~\citep{HypeTKG}. 


\paragraph{Evaluation Metric.}
We use the \textit{MR} (Mean Rank), \textit{MRR} (Mean Reciprocal Rank) and \textit{Hits@K} (K=1,3,10) as our evaluation metrics. We abbreviate \textit{`Hits@K'} as \textit{`H@K'} and employ filtering settings~\citep{transe} during the evaluation to eliminate existing facts in the dataset. It is worth noting that for the query $((h,r,?),{(k_i:v_i)}_{i=1}^{m})$ in HKGs, there are two evaluation filtering settings in existing models: one that filters out facts satisfying $((h,r,?),{(k_i:v_i)}_{i=1}^{m})$ and another that filters out facts satisfying only $(h,r,?)$ in the training set. Similarly, the difference in filtering settings of TKG occurs in timestamp. In this paper, we adopt the strict filtering setting of the former. To ensure fair comparison, for HKG we utilize the results of HyperSAT~\citep{HyperSAT} with the same settings as ours. For TKG, we thoroughly review the original code of our baselines and reproduce the results of some methods.

\paragraph{Baselines.}\textit{For HKG}, we compare with NaLP~\citep{NaLP}, StarE~\citep{stare}, GRAN~\citep{gran}, tNaLP~\citep{tnalp}, HyNT~\citep{hynt}, ShrinkE~\citep{shrinke}, HAHE~\citep{hahe}, NYLON~\citep{nylon} and HyperSAT~\citep{HyperSAT}. \textit{For NKG}, QuatE~\citep{quate}, BiQUE~\citep{bique}, BiVE~\citep{bive}, NestE~\citep{neste}, HOKE~\citep{HOKE} and GRADATE~\citep{GRADATE} are chosen as baselines. \textit{For TKG}, we compare against following methods: ATiSE~\citep{ATiSE}, TGeomE+~\citep{TGeomE+}, HGE~\citep{HGE}, DuaTHP~\citep{duathp}, ECEformer~\citep{eceformer} and 5EL~\citep{5EL}.

\begin{table*}
\centering
\renewcommand{\arraystretch}{1.0} 
\resizebox{0.987\textwidth}{!}{
\begin{tabular}{lcccc|ccccccccc}
\toprule
     & \multicolumn{2}{c}{\textbf{FBH}}                     & \multicolumn{2}{c}{\textbf{DBHE}} & \multicolumn{3}{c}{\textbf{FBH}}                        & \multicolumn{3}{c}{\textbf{FBHE}}                     & \multicolumn{3}{c}{\textbf{DBHE}}                         \\ \cmidrule(lr){2-5}\cmidrule(lr){6-14}

\multicolumn{1}{l}{\textbf{Model}}                      & MRR            & H@10                      & MRR            & \multicolumn{1}{c}{H@10}     & MR            & MRR            & H@10        & MR            & MRR            & H@10        & MR            & MRR            & H@10      \\ \cmidrule(lr){2-14} 
\multirow{1}{*}{}     & \multicolumn{4}{c|}{\textit{Base link prediction}}  & \multicolumn{9}{c}{\textit{Triple prediction}} \\ \midrule
QuatE~\citep{quate}          & 0.354          & 0.581       & 0.264          & 0.440       & 145603.8      & 0.103          & 0.114        & 94684.4       & 0.101          & 0.209       & 26485.0       & 0.157          & 0.179\\
BiQUE~\citep{bique}          & 0.356          & 0.583           & 0.274          & 0.446       & 81687.5       & 0.104          & 0.115       & 61015.2       & 0.135          & 0.205        & 19079.4       & 0.163          & 0.185\\
BiVE~\citep{bive}            & 0.370          & 0.607          & 0.274          & 0.422         & 6.20          & 0.855          & 0.941        & 8.35          & 0.711          & 0.866        & 3.63          & 0.687          & 0.958 \\

NestE~\citep{neste}         & \underline{0.371}          & 0.608   & \underline{0.289}    & \underline{0.443}    & \underline{3.34}    & \underline{0.922} & \underline{0.982}  & \textbf{3.05} & \textbf{0.851}    & \textbf{0.962}   & \underline{2.07}    & \textbf{0.862}    &\underline{0.984}       \\ 
HOKE~\citep{HOKE} & -          & -          & -         & -         & 3.06          & 0.719          & 0.777        & 2.82          & 0.674          & 0.764        & 2.10          & 0.674          & 0.777 \\

GRADATE~\citep{GRADATE} & -          & -          & -         & -         & 18.15          & 0.780          & 0.871        & 26.81          & 0.603          & 0.757        & 4.72          & 0.654          & 0.916 \\
\midrule
\textbf{UniHR}  & \textbf{0.401} & \textbf{0.619} & \textbf{0.296}    & \textbf{0.448}  & \textbf{2.46} & \textbf{0.946}          & \textbf{0.993}  
      &\underline{5.20}          & \underline{0.793} & \underline{0.890}    & \textbf{1.90} & \textbf{0.862} & \textbf{0.987}       \\ \bottomrule

\end{tabular}}
\vspace{-1.5mm}
\caption{Results of base link prediction and triple prediction. Results of NKG-specific methods are taken from original papers.}\label{tab2}
\vspace{-3mm}
\end{table*}


\paragraph{Implementation details.}
All experiments are conducted on a single Nvidia 80G A800 GPU and implemented with PyTorch. For base link prediction on NKGs, we also use augmented triples from~\citep{bive} for training to ensure fairness. For triple prediction, due to the small size of training set, we conduct training based on fixed embeddings of entities obtained from the base link prediction and set $\omega_{nested\;relation}\text{=}\ 0$ to prevent overfitting. 

\subsection{Main Results}

\paragraph{\textbf{Link Prediction on HKG.}}
We compare our method with previous methods on the WD50K and WikiPeople datasets shown in Table \ref{tab1}. Among these methods, it can be seen that our proposed UniHR achieves competitive results with the state-of-the-art method HAHE and HyperSAT, which means our method effectively captures hierarchical fact information. Compared to GNN-based method StarE, we achieve improvements of 3.9 points (12.6\%) in MRR, 4.4 points (18.8\%) in Hits@1 and 3.0 points (6.6\%) in Hits@10 on WD50K. This indicates that the performance of StarE's customized GNN is limited by its inability to flexibly capture key-value pair information and hierarchical semantics.

\begin{table}
\centering
\renewcommand{\arraystretch}{1.0} 
\resizebox{0.47\textwidth}{!}{
\begin{tabular}{lcccc}
\toprule
\multirow{2}{*}{\textbf{Model}} & \multicolumn{4}{c}{\textbf{wikidata12k}}                                   \\ \cmidrule(lr){2-5} 
                    & MRR            & H@1         & H@3         & H@10        \\ \midrule
ATiSE~\citep{ATiSE}               & 0.252          & 0.148          & 0.288          & 0.462          \\
TGeomE+~\citep{TGeomE+}                 & \underline{0.333} &{0.232}    & \underline{0.361} & \textbf{0.546} \\ 
HGE$^{*}$~\citep{HGE}         & 0.290 & 0.176    & 0.323 &0.514 \\ 
ECEformer$^{*}$~\citep{eceformer}      & 0.262   & 0.159    & 0.255   &0.462 \\
DuaTHP~\citep{duathp}      & 0.304   & 0.209    & 0.331   &0.509 \\
5EL~\citep{5EL}      & 0.311   & \underline{0.237}    & 0.355   &\textbf{0.546} \\
 \midrule
\textbf{UniHR}         & \textbf{0.334}    & \textbf{0.242} & \textbf{0.368}    & \underline{0.527}          \\  \bottomrule
\end{tabular}}
\vspace{-1.5mm}
\caption{Results of link prediction on wikidata12k. Results$^{*}$ are reported by us, and others are taken from original papers.}\label{tab3}
\vspace{-3mm}
\end{table}

\paragraph{\textbf{Link Prediction on NKG.}}
From the results in Table \ref{tab2}, we can see that our proposed UniHR obtains competitive results as the first method to capture global structural information of NKGs. For base link prediction task on triple-based KGs, UniHR achieves considerable improvements. Of particular note, the MRR of FBHE increases by 8.1\%.

For triple prediction, we perform best on FBH and DBHE datasets, especially obtaining an improvement of 2.4 points in MRR on FBH, and achieve the second-best performance on FBHE, which suggests that structural information is also valuable for NKG and UniHR can effectively capture the heterogeneity of NKG to enhance node embeddings. Unlike previous methods that use customized decoders for triples, our unified approach does not.

\begin{table}
\centering
\renewcommand{\arraystretch}{1.0} 
\resizebox{0.46\textwidth}{!}{
\begin{tabular}{lcccccc}
\toprule
\multirow{2}{*}{\textbf{Variant}}               & \multicolumn{2}{c}{\textbf{FBHE \textit{(N)}}}                        & \multicolumn{2}{c}{\textbf{DB15K \textit{(H)}}}  & \multicolumn{2}{c}{\textbf{wikidata12k \textit{(T)}}}                       \\ \cmidrule(lr){2-3} \cmidrule(lr){4-5} \cmidrule(lr){6-7} 
                     & MRR       & H@10    & MRR        & H@10         & MRR       & H@10              \\ \midrule
w/o initial $\textbf{h}_f$        & 0.767          & 0.885         & 0.346        & 0.481          & 0.333      & 0.525   \\
w/o $\textbf{W}_r$       & 0.792          & 0.885       & 0.346        & 0.480          & 0.331      & 0.521   \\
w/o intra-fact MP        & 0.754          & 0.883      & 0.341          & 0.471          & 0.321       &0.515   \\
w/o $\omega_{\tau(r)}$          & 0.782              & 0.888	&0.342	&0.476	&0.328 &0.522 \\ 
w/o $\textbf{W}_{\lambda(r)}$            & 0.778              & 0.889	&0.341	&0.474	&0.327 &0.521 \\ 
w/o inter-fact MP   	&0.776	&0.887           & 0.338       & 0.468          & 0.319       &0.511   \\ \midrule
\textbf{UniHR}              & \textbf{0.793} & \textbf{0.890} & \textbf{0.348} & \textbf{0.482} & \textbf{0.334} & \textbf{0.527} \\ \bottomrule
\end{tabular}}
\vspace{-1.5mm}
\caption{Results of ablation studies on three KG types.}\label{more_ablation}
\vspace{-3mm}
\end{table}

\paragraph{\textbf{Link Prediction on TKG.}}
As shown in Table~\ref{tab3}, we achieve competitive results on wikidata12k, even surpassing TGeomE+ by 4.3\% on Hits@1 and 1.9\% on Hits@3. However, existing TKG methods (e.g, TGeomE+ and HGE with temporal-augmented triple encoding, or ECEformer with temporal-guided subgraph encoding) only focus on partial factual semantics. In contrast, our approach efficiently encodes timestamps as atomic nodes only during initialization and learns temporal information through message passing on graph structure, demonstrating that graph structure information is also beneficial for temporal knowledge graphs, highlighting the effectiveness of our UniHR.


\begin{figure*}
\centering
\includegraphics[scale=0.8]{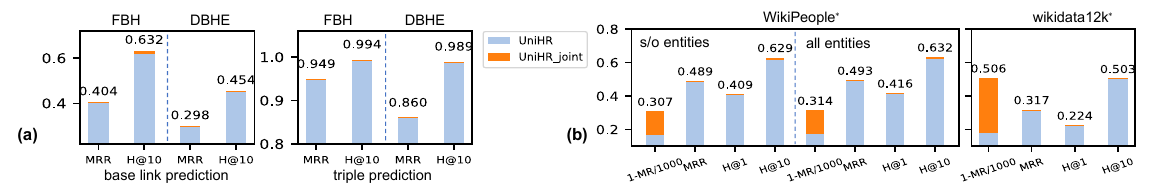}
\vspace{-4mm}
\caption{(a) improvements of joint training on hybrid tasks. (b) improvements of joint training on \textit{\textbf{wikimix}} dataset with hybrid fact forms. Yellow region indicates improvements achieved by joint training.} \label{potential}
\vspace{-2mm}
\end{figure*}

\subsection{Ablation Study on HiSL}
To analyze the contribution of different modules across various KG types, we present ablation results in Table~\ref{more_ablation}. 
It can be observed that both intra-fact and inter-fact message passing contribute to performance improvement. In particular, intra-fact message passing proves to be more beneficial for NKGs. We attribute this to fact nodes in NKGs being inherently composed of other atomic nodes, making triple prediction rely heavily on comprehensive bi-level fact semantics. In contrast, HKGs and TKGs focus solely on atomic nodes, whose representations are not dependent on other nodes. Therefore, inter-fact message passing, by capturing the global context among facts, works more effective for HKGs and TKGs, leading to better performances.

\subsection{Potential of Unified Representation}
\paragraph{\textbf{Generalize to Compositional KGs.}}
Owing to its unified representation, UniHR can flexibly generalize to compositional knowledge graphs, such as hyper-relational temporal KGs (HTKGs)~\citep{HypeTKG}, which integrate the characteristics of both HKGs and TKGs. In HTKGs, each hyper-relational fact is associated with a timestamp that explicitly indicates its temporal validity. As shown in Table \ref{tab6}, UniHR offers a performance improvement in link prediction tasks on HTKGs, outperforming both TKG-specific and HKG-specific models. This result illustrates the strong ability of UniHR to jointly model auxiliary key-value pairs and temporal information. Furthermore, UniHR achieves competitive performance with the specialized model HypeTKG, despite not relying on complex module stacking.

\begin{table}
\centering
\renewcommand{\arraystretch}{1.0} 
\resizebox{0.46\textwidth}{!}{
\begin{tabular}{lcccccccc}
\toprule
\multirow{2}{*}{\textbf{Model}} & \multicolumn{4}{c}{\textbf{WiKi-hy}} & \multicolumn{4}{c}{\textbf{YAGO-hy}}                                     \\ \cmidrule(lr){2-5} \cmidrule(lr){6-9}
                    & MRR            & H@1         & H@3         & H@10  & MRR            & H@1         & H@3         & H@10      \\ \midrule
HGE                & 0.602          & 0.507          & 0.666              & 0.765  &0.790  &0.760    &0.814    &0.837        \\
StarE          & 0.565          & 0.491          & 0.599      & 0.703    &0.765   &0.737   &0.776   &0.820          \\
GRAN               & 0.661          & 0.610          & 0.679          & 0.750    &0.808    &0.789     &0.817      &0.842    \\
HyNT              & 0.537          & 0.444          & 0.587          & 0.723     &0.763     &0.724     &0.787    &0.836          \\
HypeTKG     &\underline{0.687}    &\textbf{0.633}   &\underline{0.710}    &\underline{0.789}   &\underline{0.832}    &\textbf{0.817}    &\underline{0.838}    &\underline{0.857} \\ \midrule 
\textbf{UniHR} &\textbf{0.692}   &\underline{0.626}   &\textbf{0.716}  &\textbf{0.792}   
&\textbf{0.841}  &\underline{0.810}  &\textbf{0.841}  
&\textbf{0.862} \\ \bottomrule
\end{tabular}}
\vspace{-1.5mm}
\caption{Results on hyper-relational TKG datasets.}\label{tab6}
\vspace{-4mm}
\end{table}

\paragraph{\textbf{Joint Learning on Different Tasks of KGs.}}For link prediction on NKGs, the two subtasks, namely base link prediction and triple prediction, share the same KG during the message-passing phase under our unified representation form. Therefore, we attempt joint training on two tasks using the NKG dataset, as shown in Fig~\ref{potential}(b). Consistent with previous studies~\citep{GRADATE}, we also observe that results of joint training are generally superior to those of separate training, further confirming that nested and atomic facts can mutually enhance and complement each other's semantics.

\paragraph{\textbf{Joint Learning on Different Types of KGs.}}
We believe unified representation is key to develop pre-trained models that integrate multiple KG types. To explore this potential, we jointly train on different KG types. Notably, real-world KGs like Wikidata~\citep{wikidata} naturally contain diverse fact types. Thus, we construct a hybrid dataset \textbf{\textit{wikimix}}, by filtering two Wikidata subsets: HKG WikiPeople and TKG wikidata12k, which share 3,546 entities and 18 relations but have no overlapping facts. To prevent data leakage, we remove test entries whose main triples appear in the other subset's train set (537 from wikidata12k and 384 from WikiPeople).

\noindent\textbf{{Quantitative Analysis.}} As shown in Figure~\ref{potential}(a), we find joint learning outperforms separate learning across most metrics. Notably, MR improves by 17.1\% on HKG and 39.7\% on TKG, indicating that leveraging richer structural interactions across different fact types facilitates more effective representation learning. 

\begin{figure}
\vspace{-3mm}
\centering
\includegraphics[scale=0.48]{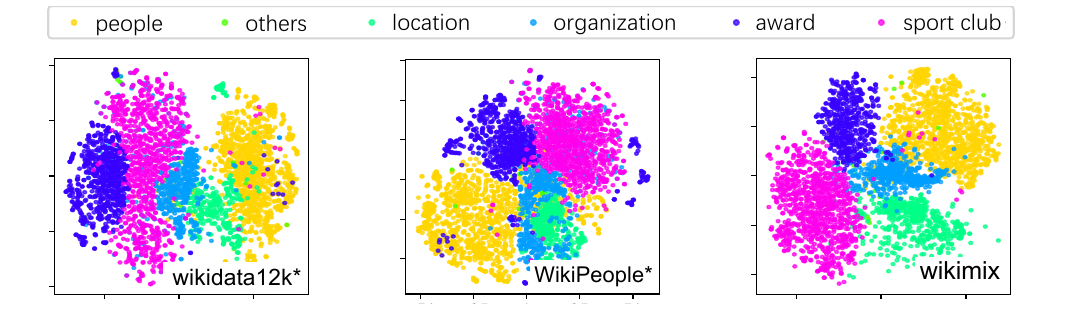}
\vspace{-8mm}
\caption{t-SNE visualization of shared entity's embeddings.} \label{visual}
\end{figure}

\noindent\textbf{{Visualization Analysis.}} As shown in Figure~\ref{visual}, entity embeddings from different categories are more coherently clustered and better separated under joint training compared to separate training, demonstrating that joint learning enables the model to acquire more structured and discriminative representations across diverse fact types.


\begin{figure}
\vspace{-3mm}
\centering
\includegraphics[scale=0.82]{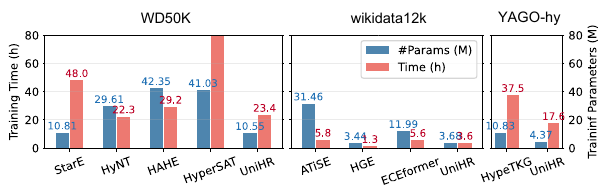}
\vspace{-8mm}
\caption{Results of efficiency analysis.} \label{effi}
\vspace{-3mm}
\end{figure}



\subsection{Efficiency Analysis}
For memory usage, HiDR as a data preprocessing module, incurs minimal additional storage overhead. Although some extra nodes and relations are introduced, only embeddings for three “connected relations” need to be stored, while embeddings for other nodes can be derived from existing atomic elements. For runtime efficiency, as shown in Figure~\ref{effi}, UniHR does not significantly increase the number of model parameters or runtime compared to state-of-the-art methods. The embeddings of newly introduced nodes are computed from atomic elements, thus avoiding parameter inflation. During message passing, we employ subgraph sampling instead of using the entire graph, and apply dropout to prevent overfitting, which effectively improves training efficiency. Overall, UniHR achieves a better trade-off between effectiveness and efficiency.

\section{Conclusion}
In this paper, we propose UniHR, a unified hierarchical KG representation learning framework consisting of a learning-optimized Hierarchical Data Representation (HiDR) module and a Hierarchical Structure Learning (HiSL) module. The HiDR module unifies hyper-relational, nested and temporal facts into the triple form. Moreover, HiSL captures local semantic information within facts and global structural information between facts. Extensive experiments show UniHR achieves the best or competitive performance across 5 types of KGs over 9 datasets and further highlight the strong potential of unified representations across 3 complex scenarios. 


\section{Acknowledgements}
This work is founded by National Natural Science Foundation of China (NSFC62306276/NSFCU23B2055), Yongjiang Talent Introduction Programme (2022A-238-G), and Fundamental Research Funds for the Central Universities (226-2023-00138). This work was supported by Ant Group.

\bibliography{aaai2026}

@inproceedings{zhanghe,
  title={Dynamic graph unlearning: a general and efficient post-processing method via gradient transformation},
  author={Zhang, He and Wu, Bang and Yang, Xiangwen and Yuan, Xingliang and Liu, Xiaoning and Yi, Xun},
  booktitle={Proceedings of the ACM on Web Conference 2025},
  pages={931--944},
  year={2025}
}

@article{prgb,
  title={Prgb benchmark: A robust placeholder-assisted algorithm for benchmarking retrieval-augmented generation},
  author={Tan, Zhehao and Jiao, Yihan and Yang, Dan and Liu, Lei and Feng, Jie and Sun, Duolin and Shen, Yue and Wang, Jian and Wei, Peng and Gu, Jinjie},
  journal={arXiv preprint arXiv:2507.22927},
  year={2025}
}

@inproceedings{eog,
  title={Enrich-on-Graph: Query-Graph Alignment for Complex Reasoning with LLM Enriching},
  author={Li, Songze and Liu, Zhiqiang and Gui, Zhengke and Chen, Huajun and Zhang, Wen},
  booktitle={Proceedings of the 2025 Conference on Empirical Methods in Natural Language Processing},
  pages={7683--7703},
  year={2025}
}

@inproceedings{rtqa,
  title={RTQA: Recursive Thinking for Complex Temporal Knowledge Graph Question Answering with Large Language Models},
  author={Gong, Zhaoyan and Li, Juan and Liu, Zhiqiang and Liang, Lei and Chen, Huajun and Zhang, Wen},
  booktitle={Proceedings of the 2025 Conference on Empirical Methods in Natural Language Processing},
  pages={9864--9881},
  year={2025}
}

@inproceedings{hitter,
  author       = {Sanxing Chen and
                  Xiaodong Liu and
                  Jianfeng Gao and
                  Jian Jiao and
                  Ruofei Zhang and
                  Yangfeng Ji},
  editor       = {Marie{-}Francine Moens and
                  Xuanjing Huang and
                  Lucia Specia and
                  Scott Wen{-}tau Yih},
  title        = {HittER: Hierarchical Transformers for Knowledge Graph Embeddings},
  booktitle    = {Proceedings of the 2021 Conference on Empirical Methods in Natural
                  Language Processing, {EMNLP} 2021, Virtual Event / Punta Cana, Dominican
                  Republic, 7-11 November, 2021},
  pages        = {10395--10407},
  publisher    = {Association for Computational Linguistics},
  year         = {2021},
  url          = {https://doi.org/10.18653/v1/2021.emnlp-main.812},
  doi          = {10.18653/V1/2021.EMNLP-MAIN.812},
  bibsource    = {dblp computer science bibliography, https://dblp.org}
}

@article{duathp,
  title={Integrating Transformer Architecture and Householder Transformations for Enhanced Temporal Knowledge Graph Embedding in DuaTHP},
  author={Chen, Yutong and Li, Xia and Liu, Yang and Hu, Tiangui},
  journal={Symmetry},
  volume={17},
  number={2},
  pages={173},
  year={2025},
  publisher={MDPI}
}

@inproceedings{5EL,
    title={Integrating Large Language Models and M{\"o}bius Group Transformations for Temporal Knowledge Graph Embedding on the Riemann Sphere},
  author={Zhang, Sensen and Liang, Xun and Niu, Simin and Niu, Zhendong and Wu, Bo and Hua, Gengxin and Wang, Long and Guan, Zhenyu and Wang, Hanyu and Zhang, Xuan and others},
  booktitle={Proceedings of the AAAI Conference on Artificial Intelligence},
  volume={39},
  pages={13277--13285},
  year={2025}
}

@inproceedings{nylon,
  author       = {Weijian Yu and
                  Jie Yang and
                  Dingqi Yang},
  editor       = {Tat{-}Seng Chua and
                  Chong{-}Wah Ngo and
                  Ravi Kumar and
                  Hady W. Lauw and
                  Roy Ka{-}Wei Lee},
  title        = {Robust Link Prediction over Noisy Hyper-Relational Knowledge Graphs
                  via Active Learning},
  booktitle    = {Proceedings of the {ACM} on Web Conference 2024, {WWW} 2024, Singapore,
                  May 13-17, 2024},
  pages        = {2282--2293},
  publisher    = {{ACM}},
  year         = {2024},
  url          = {https://doi.org/10.1145/3589334.3645686},
  doi          = {10.1145/3589334.3645686},
  bibsource    = {dblp computer science bibliography, https://dblp.org}
}

@article{shrinke,
  title={Shrinking embeddings for hyper-relational knowledge graphs},
  author={Xiong, Bo and Nayyer, Mojtaba and Pan, Shirui and Staab, Steffen},
  journal={arXiv preprint arXiv:2306.02199},
  year={2023}
}

@article{tnalp,
  author       = {Saiping Guan and
                  Xiaolong Jin and
                  Jiafeng Guo and
                  Yuanzhuo Wang and
                  Xueqi Cheng},
  title        = {Link Prediction on N-ary Relational Data Based on Relatedness Evaluation},
  journal      = {{IEEE} Trans. Knowl. Data Eng.},
  volume       = {35},
  number       = {1},
  pages        = {672--685},
  year         = {2023},
  url          = {https://doi.org/10.1109/TKDE.2021.3073483},
  doi          = {10.1109/TKDE.2021.3073483},
  bibsource    = {dblp computer science bibliography, https://dblp.org}
}

@article{HyperSAT,
  author       = {Junjie Wang and
                  Huajun Chen and
                  Wen Zhang},
  title        = {Structure-Aware Transformer for hyper-relational knowledge graph completion},
  journal      = {Expert Syst. Appl.},
  volume       = {277},
  pages        = {126992},
  year         = {2025},
  url          = {https://doi.org/10.1016/j.eswa.2025.126992},
  doi          = {10.1016/J.ESWA.2025.126992},
  bibsource    = {dblp computer science bibliography, https://dblp.org}
}

@inproceedings{HOKE,
  author       = {Giuseppe Pirr{\`{o}}},
  editor       = {Claudia Hauff and
                  Craig Macdonald and
                  Dietmar Jannach and
                  Gabriella Kazai and
                  Franco Maria Nardini and
                  Fabio Pinelli and
                  Fabrizio Silvestri and
                  Nicola Tonellotto},
  title        = {Higher Order Knowledge Graph Embeddings},
  booktitle    = {Advances in Information Retrieval - 47th European Conference on Information
                  Retrieval, {ECIR} 2025, Lucca, Italy, April 6-10, 2025, Proceedings,
                  Part {I}},
  series       = {Lecture Notes in Computer Science},
  volume       = {15572},
  pages        = {181--195},
  publisher    = {Springer},
  year         = {2025},
  url          = {https://doi.org/10.1007/978-3-031-88708-6\_12},
  doi          = {10.1007/978-3-031-88708-6\_12},
  bibsource    = {dblp computer science bibliography, https://dblp.org}
}

@article{GRADATE,
  author       = {Hao Li and
                  Ke Liang and
                  Wenjing Yang and
                  Lingyuan Meng and
                  Yaohua Wang and
                  Sihang Zhou and
                  Xinwang Liu},
  title        = {Eyes on Islanded Nodes: Better Reasoning via Structure Augmentation
                  and Feature Co-Training on Bi-Level Knowledge Graphs},
  journal      = {{IEEE} Trans. Image Process.},
  volume       = {34},
  pages        = {3268--3280},
  year         = {2025},
  url          = {https://doi.org/10.1109/TIP.2025.3572825},
  doi          = {10.1109/TIP.2025.3572825},
  bibsource    = {dblp computer science bibliography, https://dblp.org}
}

@inproceedings{HypeTKG,
  author       = {Zifeng Ding and
                  Jingcheng Wu and
                  Jingpei Wu and
                  Yan Xia and
                  Bo Xiong and
                  Volker Tresp},
  editor       = {Yaser Al{-}Onaizan and
                  Mohit Bansal and
                  Yun{-}Nung Chen},
  title        = {Temporal Fact Reasoning over Hyper-Relational Knowledge Graphs},
  booktitle    = {Findings of the Association for Computational Linguistics: {EMNLP}
                  2024, Miami, Florida, USA, November 12-16, 2024},
  pages        = {355--373},
  publisher    = {Association for Computational Linguistics},
  year         = {2024},
  url          = {https://aclanthology.org/2024.findings-emnlp.20},
  bibsource    = {dblp computer science bibliography, https://dblp.org}
}

@inproceedings{HGE,
  title={HGE: embedding temporal knowledge graphs in a product space of heterogeneous geometric subspaces},
  author={Pan, Jiaxin and Nayyeri, Mojtaba and Li, Yinan and Staab, Steffen},
  booktitle={Proceedings of the AAAI Conference on Artificial Intelligence},
  volume={38},
  pages={8913--8920},
  year={2024}
}

@article{Time2Vec,
  author       = {Seyed Mehran Kazemi and
                  Rishab Goel and
                  Sepehr Eghbali and
                  Janahan Ramanan and
                  Jaspreet Sahota and
                  Sanjay Thakur and
                  Stella Wu and
                  Cathal Smyth and
                  Pascal Poupart and
                  Marcus A. Brubaker},
  title        = {Time2Vec: Learning a Vector Representation of Time},
  journal      = {CoRR},
  volume       = {abs/1907.05321},
  year         = {2019},
  url          = {http://arxiv.org/abs/1907.05321},
  eprinttype    = {arXiv},
  eprint       = {1907.05321},
  bibsource    = {dblp computer science bibliography, https://dblp.org}
}

@article{RDF,
  author       = {Waqas Ali and
                  Muhammad Saleem and
                  Bin Yao and
                  Aidan Hogan and
                  Axel{-}Cyrille Ngonga Ngomo},
  title        = {A survey of {RDF} stores {\&} {SPARQL} engines for querying knowledge
                  graphs},
  journal      = {{VLDB} J.},
  volume       = {31},
  number       = {3},
  pages        = {1--26},
  year         = {2022},
  url          = {https://doi.org/10.1007/s00778-021-00711-3},
  doi          = {10.1007/S00778-021-00711-3},
  bibsource    = {dblp computer science bibliography, https://dblp.org}
}

@article{TGeomE+,
  author       = {Chengjin Xu and
                  Mojtaba Nayyeri and
                  Yung{-}Yu Chen and
                  Jens Lehmann},
  title        = {Geometric Algebra Based Embeddings for Static and Temporal Knowledge
                  Graph Completion},
  journal      = {{IEEE} Trans. Knowl. Data Eng.},
  volume       = {35},
  number       = {5},
  pages        = {4838--4851},
  year         = {2023},
  url          = {https://doi.org/10.1109/TKDE.2022.3151435},
  doi          = {10.1109/TKDE.2022.3151435},
  bibsource    = {dblp computer science bibliography, https://dblp.org}
}

@inproceedings{beyond_triple,
  title={Reasoning beyond triples: Recent advances in knowledge graph embeddings},
  author={Xiong, Bo and Nayyeri, Mojtaba and Daza, Daniel and Cochez, Michael},
  booktitle={Proceedings of the 32nd ACM International Conference on Information and Knowledge Management},
  pages={5228--5231},
  year={2023}
}

@inproceedings{eceformer,
  author       = {Zhiyu Fang and
                  Shuai{-}Long Lei and
                  Xiaobin Zhu and
                  Chun Yang and
                  Shi{-}Xue Zhang and
                  Xu{-}Cheng Yin and
                  Jingyan Qin},
  editor       = {Grace Hui Yang and
                  Hongning Wang and
                  Sam Han and
                  Claudia Hauff and
                  Guido Zuccon and
                  Yi Zhang},
  title        = {Transformer-based Reasoning for Learning Evolutionary Chain of Events
                  on Temporal Knowledge Graph},
  booktitle    = {Proceedings of the 47th International {ACM} {SIGIR} Conference on
                  Research and Development in Information Retrieval, {SIGIR} 2024, Washington
                  DC, USA, July 14-18, 2024},
  pages        = {70--79},
  publisher    = {{ACM}},
  year         = {2024},
  url          = {https://doi.org/10.1145/3626772.3657706},
  doi          = {10.1145/3626772.3657706},
  bibsource    = {dblp computer science bibliography, https://dblp.org}
}

@inproceedings{hahe,
  author       = {Haoran Luo and
                  Haihong E and
                  Yuhao Yang and
                  Yikai Guo and
                  Mingzhi Sun and
                  Tianyu Yao and
                  Zichen Tang and
                  Kaiyang Wan and
                  Meina Song and
                  Wei Lin},
  editor       = {Anna Rogers and
                  Jordan L. Boyd{-}Graber and
                  Naoaki Okazaki},
  title        = {{HAHE:} Hierarchical Attention for Hyper-Relational Knowledge Graphs
                  in Global and Local Level},
  booktitle    = {Proceedings of the 61st Annual Meeting of the Association for Computational
                  Linguistics (Volume 1: Long Papers), {ACL} 2023, Toronto, Canada,
                  July 9-14, 2023},
  pages        = {8095--8107},
  publisher    = {Association for Computational Linguistics},
  year         = {2023},
  url          = {https://doi.org/10.18653/v1/2023.acl-long.450},
  doi          = {10.18653/V1/2023.ACL-LONG.450},
  bibsource    = {dblp computer science bibliography, https://dblp.org}
}

@inproceedings{NaLP,
  title={Link prediction on n-ary relational data},
  author={Guan, Saiping and Jin, Xiaolong and Wang, Yuanzhuo and Cheng, Xueqi},
  booktitle={Proceedings of the 28th International Conference on World Wide Web (WWW'19)},
  year={2019},
  pages={583--593}
}

@inproceedings{stare,
  title={Message Passing for Hyper-Relational Knowledge Graphs},
  author={Galkin, Mikhail and Trivedi, Priyansh and Maheshwari, Gaurav and Usbeck, Ricardo and Lehmann, Jens},
  booktitle={Proceedings of the 2020 Conference on Empirical Methods in Natural Language Processing (EMNLP)},
  pages={7346--7359},
  year={2020}
}

@inproceedings{gran,
  title={Link Prediction on N-ary Relational Facts: A Graph-based Approach},
  author={Wang, Quan and Wang, Haifeng and Lyu, Yajuan and Zhu, Yong},
  booktitle={Findings of the Association for Computational Linguistics: ACL-IJCNLP 2021},
  pages={396--407},
  year={2021}
}

@inproceedings{hynt,
  author       = {Chanyoung Chung and
                  Jaejun Lee and
                  Joyce Jiyoung Whang},
  editor       = {Ambuj K. Singh and
                  Yizhou Sun and
                  Leman Akoglu and
                  Dimitrios Gunopulos and
                  Xifeng Yan and
                  Ravi Kumar and
                  Fatma Ozcan and
                  Jieping Ye},
  title        = {Representation Learning on Hyper-Relational and Numeric Knowledge
                  Graphs with Transformers},
  booktitle    = {Proceedings of the 29th {ACM} {SIGKDD} Conference on Knowledge Discovery
                  and Data Mining, {KDD} 2023, Long Beach, CA, USA, August 6-10, 2023},
  pages        = {310--322},
  publisher    = {{ACM}},
  year         = {2023},
  url          = {https://doi.org/10.1145/3580305.3599490},
  doi          = {10.1145/3580305.3599490},
  bibsource    = {dblp computer science bibliography, https://dblp.org}
}

@inproceedings{CompGCN,
  title={Composition-based Multi-Relational Graph Convolutional Networks},
  author={Vashishth, Shikhar and Sanyal, Soumya and Nitin, Vikram and Talukdar, Partha},
  booktitle={International Conference on Learning Representations},
  year={2019}
}

@inproceedings{ma-gnn,
  title={Double-Branch Multi-Attention based Graph Neural Network for Knowledge Graph Completion},
  author={Xu, Hongcai and Bao, Junpeng and Liu, Wenbo},
  booktitle={Proceedings of the 61st Annual Meeting of the Association for Computational Linguistics (Volume 1: Long Papers)},
  pages={15257--15271},
  year={2023}
}

@article{transformer,
  title={Attention is all you need},
  author={Vaswani, Ashish and Shazeer, Noam and Parmar, Niki and Uszkoreit, Jakob and Jones, Llion and Gomez, Aidan N and Kaiser, {\L}ukasz and Polosukhin, Illia},
  journal={Advances in neural information processing systems},
  volume={30},
  year={2017}
}

@article{quate,
  title={Quaternion knowledge graph embeddings},
  author={Zhang, Shuai and Tay, Yi and Yao, Lina and Liu, Qi},
  journal={Advances in neural information processing systems},
  volume={32},
  year={2019}
}

@inproceedings{transe,
  author       = {Antoine Bordes and
                  Nicolas Usunier and
                  Alberto Garc{\'{\i}}a{-}Dur{\'{a}}n and
                  Jason Weston and
                  Oksana Yakhnenko},
  editor       = {Christopher J. C. Burges and
                  L{\'{e}}on Bottou and
                  Zoubin Ghahramani and
                  Kilian Q. Weinberger},
  title        = {Translating Embeddings for Modeling Multi-relational Data},
  booktitle    = {Advances in Neural Information Processing Systems 26: 27th Annual
                  Conference on Neural Information Processing Systems 2013. Proceedings
                  of a meeting held December 5-8, 2013, Lake Tahoe, Nevada, United States},
  pages        = {2787--2795},
  year         = {2013},
  url          = {https://proceedings.neurips.cc/paper/2013/hash/1cecc7a77928ca8133fa24680a88d2f9-Abstract.html},
  bibsource    = {dblp computer science bibliography, https://dblp.org}
}

@inproceedings{bique,
  title={BiQUE: Biquaternionic Embeddings of Knowledge Graphs},
  author={Guo, Jia and Kok, Stanley},
  booktitle={Proceedings of the 2021 Conference on Empirical Methods in Natural Language Processing},
  pages={8338--8351},
  year={2021}
}

@inproceedings{bive,
  author       = {Chanyoung Chung and
                  Joyce Jiyoung Whang},
  editor       = {Brian Williams and
                  Yiling Chen and
                  Jennifer Neville},
  title        = {Learning Representations of Bi-level Knowledge Graphs for Reasoning
                  beyond Link Prediction},
  booktitle    = {Thirty-Seventh {AAAI} Conference on Artificial Intelligence, {AAAI}
                  2023, Thirty-Fifth Conference on Innovative Applications of Artificial
                  Intelligence, {IAAI} 2023, Thirteenth Symposium on Educational Advances
                  in Artificial Intelligence, {EAAI} 2023, Washington, DC, USA, February
                  7-14, 2023},
  pages        = {4208--4216},
  publisher    = {{AAAI} Press},
  year         = {2023},
  url          = {https://doi.org/10.1609/aaai.v37i4.25538},
  doi          = {10.1609/AAAI.V37I4.25538},
  bibsource    = {dblp computer science bibliography, https://dblp.org}
}

@inproceedings{neste,
  title={NestE: modeling nested relational structures for knowledge graph reasoning},
  author={Xiong, Bo and Nayyeri, Mojtaba and Luo, Linhao and Wang, Zihao and Pan, Shirui and Staab, Steffen},
  booktitle={Proceedings of the AAAI Conference on Artificial Intelligence},
  volume={38},
  pages={9205--9213},
  year={2024}
}

@inproceedings{rotate,
  title={RotatE: Knowledge Graph Embedding by Relational Rotation in Complex Space},
  author={Sun, Zhiqing and Deng, Zhi-Hong and Nie, Jian-Yun and Tang, Jian},
  booktitle={International Conference on Learning Representations},
  year={2018}
}

@inproceedings{conve,
  author       = {Tim Dettmers and
                  Pasquale Minervini and
                  Pontus Stenetorp and
                  Sebastian Riedel},
  editor       = {Sheila A. McIlraith and
                  Kilian Q. Weinberger},
  title        = {Convolutional 2D Knowledge Graph Embeddings},
  booktitle    = {Proceedings of the Thirty-Second {AAAI} Conference on Artificial Intelligence,
                  (AAAI-18), the 30th innovative Applications of Artificial Intelligence
                  (IAAI-18), and the 8th {AAAI} Symposium on Educational Advances in
                  Artificial Intelligence (EAAI-18), New Orleans, Louisiana, USA, February
                  2-7, 2018},
  pages        = {1811--1818},
  publisher    = {{AAAI} Press},
  year         = {2018},
  url          = {https://doi.org/10.1609/aaai.v32i1.11573},
  doi          = {10.1609/AAAI.V32I1.11573},
  bibsource    = {dblp computer science bibliography, https://dblp.org}
}

@inproceedings{hyte,
  title={Hyte: Hyperplane-based temporally aware knowledge graph embedding},
  author={Dasgupta, Shib Sankar and Ray, Swayambhu Nath and Talukdar, Partha},
  booktitle={Proceedings of the 2018 conference on empirical methods in natural language processing},
  pages={2001--2011},
  year={2018}
}

@article{ATiSE,
  title={Temporal knowledge graph embedding model based on additive time series decomposition},
  author={Xu, Chengjin and Nayyeri, Mojtaba and Alkhoury, Fouad and Yazdi, Hamed Shariat and Lehmann, Jens},
  journal={arXiv preprint arXiv:1911.07893},
  year={2019}
}

@article{wikidata,
  title={Wikidata: a free collaborative knowledgebase},
  author={Vrande{\v{c}}i{\'c}, Denny and Kr{\"o}tzsch, Markus},
  journal={Communications of the ACM},
  volume={57},
  number={10},
  pages={78--85},
  year={2014},
  publisher={ACM New York, NY, USA}
}

@inproceedings{bollacker2008freebase,
  title={Freebase: a collaboratively created graph database for structuring human knowledge},
  author={Bollacker, Kurt and Evans, Colin and Paritosh, Praveen and Sturge, Tim and Taylor, Jamie},
  booktitle={Proceedings of the 2008 ACM SIGMOD international conference on Management of data},
  pages={1247--1250},
  year={2008}
}

@inproceedings{kaiser2021reinforcement,
  title={Reinforcement learning from reformulations in conversational question answering over knowledge graphs},
  author={Kaiser, Magdalena and Saha Roy, Rishiraj and Weikum, Gerhard},
  booktitle={Proceedings of the 44th international ACM SIGIR conference on research and development in information retrieval},
  pages={459--469},
  year={2021}
}

@inproceedings{annervaz2018learning,
  title={Learning beyond Datasets: Knowledge Graph Augmented Neural Networks for Natural Language Processing},
  author={Annervaz, KM and Chowdhury, Somnath Basu Roy and Dukkipati, Ambedkar},
  booktitle={Proceedings of the 2018 Conference of the North American Chapter of the Association for Computational Linguistics: Human Language Technologies, Volume 1 (Long Papers)},
  pages={313--322},
  year={2018}
}

@inproceedings{trustuqa,
  title={Trustuqa: A trustful framework for unified structured data question answering},
  author={Zhang, Wen and Jin, Long and Zhu, Yushan and Chen, Jiaoyan and Huang, Zhiwei and Wang, Junjie and Hua, Yin and Liang, Lei and Chen, Huajun},
  booktitle={Proceedings of the AAAI Conference on Artificial Intelligence},
  volume={39},
  pages={25931--25939},
  year={2025}
}

@article{DBpedia,
  author       = {Jens Lehmann and
                  Robert Isele and
                  Max Jakob and
                  Anja Jentzsch and
                  Dimitris Kontokostas and
                  Pablo N. Mendes and
                  Sebastian Hellmann and
                  Mohamed Morsey and
                  Patrick van Kleef and
                  S{\"{o}}ren Auer and
                  Christian Bizer},
  title        = {DBpedia - {A} large-scale, multilingual knowledge base extracted from
                  Wikipedia},
  journal      = {Semantic Web},
  volume       = {6},
  number       = {2},
  pages        = {167--195},
  year         = {2015},
  url          = {https://doi.org/10.3233/SW-140134},
  doi          = {10.3233/SW-140134},
  bibsource    = {dblp computer science bibliography, https://dblp.org}
}

@inproceedings{unified_rec,
  title={Deep unified representation for heterogeneous recommendation},
  author={Lu, Chengqiang and Yin, Mingyang and Shen, Shuheng and Ji, Luo and Liu, Qi and Yang, Hongxia},
  booktitle={Proceedings of the ACM Web Conference 2022},
  pages={2141--2152},
  year={2022}
}

@inproceedings{ontotune,
  author       = {Zhiqiang Liu and
                  Chengtao Gan and
                  Junjie Wang and
                  Yichi Zhang and
                  Zhongpu Bo and
                  Mengshu Sun and
                  Huajun Chen and
                  Wen Zhang},
  editor       = {Guodong Long and
                  Michale Blumestein and
                  Yi Chang and
                  Liane Lewin{-}Eytan and
                  Zi Helen Huang and
                  Elad Yom{-}Tov},
  title        = {OntoTune: Ontology-Driven Self-training for Aligning Large Language
                  Models},
  booktitle    = {Proceedings of the {ACM} on Web Conference 2025, {WWW} 2025, Sydney,
                  NSW, Australia, 28 April 2025- 2 May 2025},
  pages        = {119--133},
  publisher    = {{ACM}},
  year         = {2025},
  url          = {https://doi.org/10.1145/3696410.3714816},
  doi          = {10.1145/3696410.3714816},
  bibsource    = {dblp computer science bibliography, https://dblp.org}
}

@article{ska-bench,
  author       = {Zhiqiang Liu and
                  Enpei Niu and
                  Yin Hua and
                  Mengshu Sun and
                  Lei Liang and
                  Huajun Chen and
                  Wen Zhang},
  title        = {SKA-Bench: {A} Fine-Grained Benchmark for Evaluating Structured Knowledge
                  Understanding of LLMs},
  journal      = {CoRR},
  volume       = {abs/2507.17178},
  year         = {2025},
  url          = {https://doi.org/10.48550/arXiv.2507.17178},
  doi          = {10.48550/ARXIV.2507.17178},
  eprinttype    = {arXiv},
  eprint       = {2507.17178},
  bibsource    = {dblp computer science bibliography, https://dblp.org}
}



\appendix

\section{Appendix}

\section{A Decoder Analysis}\label{app:decoder}
To explore the effectiveness of our UniHR encoding further, we pair UniHR with different decoders and evaluated them on triple prediction task. In addition to the previously mentioned unified framework \textbf{UniHR + Transformer}, we also experiment on \textbf{UniHR + ConvE} with two scoring strategies. The ConvE \cite{conve} is the decoder customized for triples and its scoring function is $vec\left( \sigma \left( \left[ \mathbf{\tilde{h}}_h;\mathbf{\tilde{e}}_r\right] *\psi \right) \right)$, where $\mathbf{\tilde{h}}_h$ and $\mathbf{\tilde{e}}_r$ represent reshaped 2D embeddings of head entity $h$ and relation $r$, and $*$ is a convolution operator. The $vec\left( \cdot \right)$ and $\psi$ are denoted as the vectorization function and a set of convolution kernels.

\begin{table}[ht]
\centering
\renewcommand{\arraystretch}{1.0} 
\resizebox{0.4\textwidth}{!}{
\begin{tabular}{lcccccc}
\toprule
\multirow{2}{*}{\textbf{Model}}     & \multicolumn{2}{c}{\textbf{FBHE/FBH}}                    & \multicolumn{2}{c}{\textbf{DBHE}}                         \\ \cmidrule(lr){2-3} \cmidrule(lr){4-5} 
                     & MRR            & Hits@10           & MRR            & Hits@10         \\ \midrule
QuatE       & 0.354          & 0.581           & 0.264          & 0.440          \\
BiQUE       & 0.356          & 0.583           & 0.274          & 0.446          \\
BiVE       & 0.370          & 0.607           & 0.274          & 0.422          \\
NestE      & 0.371          & \underline{0.608}   & 0.289    & 0.443          \\ \midrule
{UniHR + ConvE $s_h$}  & \underline{0.397} & \textbf{0.622}      & 0.289    & 0.443          \\
{UniHR + ConvE $s_f$}   & 0.375    & 0.596               & \textbf{0.307} & \textbf{0.471} \\ 
{UniHR + Transformer} & \textbf{0.401} & \underline{0.619}           & \underline{0.296}    & \underline{0.448}          \\\bottomrule
\end{tabular}}
\caption{Base link prediction on FBHE, FBH and DBHE. All baselines' results are taken from \cite{neste}. The best results among all models are written bold, while the second are underlined. The $s_f$ and $s_h$ denote $\left(f,has\;head\;entity,h\right)\ \left(f,has\;tail\;entity,t\right)$ and $\left(h,r,t\right)$ two types of scoring method respectively.}\label{tabb1}
\end{table}

Due to our special representation, there exists two scoring methods for atomic triples, thus we present the base link prediction results separately for each scoring method. The $s_f$ represents scoring triples $\left(f,has\;head\;entity,h\right)\ $and $\left(f,has\;tail\;entity,t\right)$, and $s_t$ represents scoring $\left(h, r, t\right)$. The performance of base link prediction is shown in Table \ref{tabb1}. It can be observed that regardless of the scoring method employed, we both achieve competitive performance, especially with scoring $\left(h, r, t\right)$ on FBH and scoring $\left(f,has\;head\;entity,h\right)\, \left(f,has\;tail\;entity,t\right)$ on DBHE. We attribute the differences in performance under different scoring methods to dataset characteristics. DBHE dataset is relatively smaller, and scoring method $s_f$ effectively alleviates overfitting problem. Conversely, for larger datasets FBH, scoring based on $\left(h, r, t\right)$ minimizes information loss.

Table \ref{tabb2} shows the results of triple prediction on three benchmark datasets. Among all baselines, Quate and Bique struggle to model the mapping relationship between atomic facts and nested facts. Furthermore, prior works \cite{bive} do not guarantee that all atomic facts in the nested fact test set are present in the training set as entities, which shifts the problem from a transductive setting to an inductive setting, leading to significant performance gaps between these baselines. On most metrics, our method outperforms BiVE and NestE which are specifically modeled for nested facts. Notably, NestE fully preserves the semantics of atomic facts. However, on the FBHE dataset, UniHR + ConvE achieves an improvement of 0.58 (6.4\%) points in MRR and 0.24 (2.4\%) points in Hits@10 compared to the state-of-the-art model NestE and the second-best performance after UniHR + Transformer on the FBH and DBHE datasets, demonstrating UniHR's powerful graph structure encoding capabilities. We also carry out ablation experiments on UniHR + ConvE as shown in Table \ref{tabb2}. Performance declines are observed after removing any part of the HiSL module, showing the significance of HiSL for hierarchical encoding. 


\begin{table*}[ht]
\centering
\renewcommand{\arraystretch}{1.0} 
\resizebox{0.75\textwidth}{!}{
\begin{tabular}{lccccccccc}
\toprule
\multirow{2}{*}{\textbf{Model}}         & \multicolumn{3}{c}{\textbf{FBH}}                         & \multicolumn{3}{c}{\textbf{FBHE}}                        & \multicolumn{3}{c}{\textbf{DBHE}}                        \\ \cmidrule(lr){2-4}\cmidrule(lr){5-7}\cmidrule(lr){8-10}
              & MR            & MRR            & Hits@10         & MR            & MRR            & Hits@10         & MR            & MRR            & Hits@10         \\ \midrule
QuatE         & 145603.8      & 0.103          & 0.114          & 94684.4       & 0.101          & 0.209          & 26485.0       & 0.157          & 0.179          \\
BiQUE       & 81687.5       & 0.104          & 0.115          & 61015.2       & 0.135          & 0.205          & 19079.4       & 0.163          & 0.185          \\
BiVE       & 6.20          & 0.855          & 0.941          & 8.35          & 0.711          & 0.866          & 3.63          & 0.687          & 0.958          \\
NestE   & 3.34    & 0.922 &0.982    & \textbf{3.05} & 0.851    & 0.962    & 2.07    & 0.862    & 0.984 \\ 
HOKE        & 3.06          & 0.719          & 0.777        & 2.82          & 0.674          & 0.764        & 2.10          & 0.674          & 0.777 \\

GRADATE  & 18.15          & 0.780          & 0.871        & 26.81          & 0.603          & 0.757        & 4.72          & 0.654          & 0.916 \\ \midrule
{UniHR + Transformer} & \textbf{2.46} & \textbf{0.946}          & \textbf{0.993} & 5.20          & 0.793 & 0.890 & \textbf{1.90} & 0.862 & \textbf{0.987}          \\ 
{UniHR + ConvE} & 3.00 &0.900          & 0.983  & 6.27          & \textbf{0.909} & \textbf{0.986} & 2.06 & \textbf{0.876} & 0.978          \\ \midrule
{UniHR + ConvE} w/o $\textbf{h}_f$  & 4.39       & 0.887          & 0.979          & 10.10        & 0.865          & 0.970          & 2.76        & 0.798          & 0.961          \\
{UniHR + ConvE} w/o intra-fact    & 6.54          & 0.859          & 0.959          & 18.10          & 0.871          & 0.968          & 5.82          & 0.665          & 0.900          \\
{UniHR + ConvE} w/o inter-fact    & 12.56             & 0.864              & 0.961              & 20.56             & 0.864              & 0.966              & 10.75         & 0.764          & 0.951          \\ \bottomrule
\end{tabular}}
\vspace{-1mm}
\caption{Triple prediction on FBHE, FBH and DBHE. }\label{tabb2}
\vspace{-3mm}
\end{table*}

\begin{table*}
\centering
\renewcommand{\arraystretch}{1.0} 
\resizebox{0.9\textwidth}{!}{
\begin{tabular}{lcccccccccccc}
\toprule
\textbf{Dataset}    
& \textbf{Fact} & \textbf{Entities} &\textbf{Rela} & \textbf{Train}  & \textbf{Valid} &\textbf{Test}  & \textbf{with Q(\%)}    & \textbf{Arity} & \textbf{N-Fact} & \textbf{N-Rela} & \textbf{AF(\%)} & \textbf{Period}  \\ \midrule
\multicolumn{13}{c}{\textit{Hyper-relational Knowledge Graph}}   \\ \midrule
\multicolumn{1}{l}{\textbf{WikiPeople}} & 369866      & 34825    & 178       & 294439 & 37715 & \multicolumn{1}{c}{37712} & 9482(2.6\%)   & \multicolumn{1}{c}{2-7}   & -           & -               & \multicolumn{1}{c}{-}       & -       \\
\multicolumn{1}{l}{\textbf{WD50K}}      & 236507      & 47155    & 531       & 166435 & 23913 & \multicolumn{1}{c}{46159} & 32167(13.6\%) & \multicolumn{1}{c}{2-67}  & -           & -               & \multicolumn{1}{c}{-}       & -       \\ \midrule
\multicolumn{13}{c}{\textit{Nested Factual Knowledge Graph}}   \\ \midrule
\multicolumn{1}{l}{\textbf{FBH}}        & 310116      & 14541    & 237       & 248094 & 31011 & \multicolumn{1}{c}{31011} & -             & \multicolumn{1}{c}{-}     & 27062       & 6               & \multicolumn{1}{c}{33157}   & -       \\
\multicolumn{1}{l}{\textbf{FBHE}}       & 310116      & 14541    & 237       & 248094 & 31011 & \multicolumn{1}{c}{31011} & -             & \multicolumn{1}{c}{-}     & 34941       & 10              & \multicolumn{1}{c}{33719}   & -       \\
\multicolumn{1}{l}{\textbf{DBHE}}       & 68296       & 12440    & 87        & 54636  & 6830  & \multicolumn{1}{c}{6830}  & -             & \multicolumn{1}{c}{-}     & 6717        & 8               & \multicolumn{1}{c}{8206}    & -       \\ \midrule
\multicolumn{13}{c}{\textit{Temporal Knowledge Graph}}   \\ \midrule
\multicolumn{1}{l}{\textbf{wikidata12k}}                     & 40621       & 12554    & 24        & 32497  & 4062  & \multicolumn{1}{c}{4062}  & -             & \multicolumn{1}{c}{-}     & -           & -               & \multicolumn{1}{c}{-}       & 19-2020 \\ \midrule

\multicolumn{13}{c}{\textit{Hyper-relational Temporal Knowledge Graph}}   \\ \midrule
\multicolumn{1}{l}{\textbf{WiKi-hy}}                     & 139078       & 16634    & 147        & 111252  & 13900  & \multicolumn{1}{c}{13926}               & \multicolumn{1}{c}{13335(9.59\%)}     & 2-8       & -    & -               & \multicolumn{1}{c}{-}       & 1513-2020 \\ 
\multicolumn{1}{l}{\textbf{YAGO-hy}}                     & 73143       & 16167    & 54        & 51193  & 10973  & \multicolumn{1}{c}{10977}  & 5107(6.98\%)             & \multicolumn{1}{c}{2-5}     & -           & -               & \multicolumn{1}{c}{-}       & 0-187 \\ \midrule

\multicolumn{13}{c}{\textit{Multiple types of Knowledge Graph}}   \\ \midrule
\multicolumn{1}{l}{\textbf{wikimix}}                     & 409566       & 43832    &184 &326936        & 41777    &3525/37328  & 9098(2.2\%)             & 2-7     & -           & -               & \multicolumn{1}{c}{-}       & 19-2020 \\ \bottomrule
\end{tabular}}
\caption{The statistics of diverse knowledge graphs dataset, where ``with Q(\%)" and ``Arity" column respectively denote the number of facts with auxiliary key-value pairs and the range of arity of hyper-relational facts, ``N-Fact'' is the number of nested fact, ``N-Rela'' is the number of nested relation, the ``AF(\%)" column denotes the number of atomic facts in nested facts.}\label{tabc1}
\end{table*}


\begin{table*}[ht]
\centering
\renewcommand{\arraystretch}{1.0} 
\resizebox{0.9\textwidth}{!}{
\begin{tabular}{lccccccccc}
\toprule
\textbf{Hyperparameter} & \textbf{WikiPeople} & \textbf{WD50K} & \textbf{wikidata12k} & \textbf{FBHE$_{base}$} & \textbf{FBH$_{base}$} & \textbf{DBHE$_{base}$} &\textbf{FBHE$_{triple}$} & \textbf{FBH$_{triple}$} & \textbf{DBHE$_{triple}$}\\ \midrule
batch\_size             & 2048                & 2048           & 2048                           & 2048          & 2048         & 2048  & 2048& 2048& 2048        \\
embedding dim           & 200                 & 200                       & 200    & 200& 200& 200              & 200           & 200          & 200           \\
hidden dim              & 200    & 200& 200& 200             & 200                       & 200                  & 200           & 200          & 200           \\
GNN\_layer              & 2   & 2  & 2  & 2                  & 2                           & 2                    & 2             & 2            & 2             \\
GNN\_intra-fact heads          & 4     & 4  & 4  & 4                & 4                           & 4                    & 4             & 4            & 4             \\
transformer layers      & 2                   & 2                           & 2                    & 2             & 2     & 2  & 2  & 2         & 2             \\
transformer heads       & 4    & 4 & 4 & 4                & 4                            & 4                    & 4             & 4            & 4             \\
transfomer activation   & gelu   & gelu  & gelu  & gelu               & gelu                      & gelu                 & gelu             & gelu            & gelu             \\
decoder dropout         & 0.1                 & 0.1                       & 0.1                  & 0.1           & 0.1          & 0.1           & 0.1& 0.1& 0.1\\
soft label for entity   & 0.2                 & 0.2                      & 0.4                  & 0.2           & 0.2          & 0.3   & 0.2 & 0.2 & 0.2        \\
soft label for relation & 0.1                 & 0.1                        & 0.3                  & 0.2           & 0.2          & 0.3   & 0.2   & 0.2   & 0.2          \\
weight\_decay           & 0.01 & 0.01 & 0.01 & 0.01                & 0.01                    & 0.01                 & 0.01          & 0.01         & 0.01          \\ 
learning rate &5e-4 &5e-4 &5e-4 &5e-4 &5e-4 &5e-4 &5e-4 &5e-4 &5e-4 \\\bottomrule
\end{tabular}}
\caption{The major hyperparameters of our approach for all link prediction tasks.}\label{tabd1}
\end{table*}

\begin{table*}
\centering
\renewcommand{\arraystretch}{1.0} 
\resizebox{0.84\textwidth}{!}{
\begin{tabular}{lccccccccccccccc}
\toprule
                & \multicolumn{10}{c}{\textbf{WikiPeople$^{*}$}}                                        & \multicolumn{5}{c}{\textbf{wikidata12k$^{*}$}}    \\ \cmidrule(lr){2-11} \cmidrule(lr){12-16}
            \multicolumn{1}{l}{\textbf{Model}}                 & \multicolumn{5}{c}{subject/object} & \multicolumn{5}{c}{all entities} & \multicolumn{5}{c}{subject/object} \\ \cmidrule(lr){2-6} 
                         \cmidrule(lr){7-11}
                         \cmidrule(lr){12-16}
                         & MR      & MRR       & H@1    & H@3    & H@10    & MR     & MRR      & H@1    & H@3    & H@10   & MR  & MRR    & H@1 & H@3  & H@10 \\ \midrule
\textbf{UniHR}    &835.8	&0.486	&\textbf{0.412}	&0.528	&0.617	&829.0	&0.488	&0.414	&0.531	&0.620	&818.7	&0.314	&0.220	&0.345	&\textbf{0.509}   \\
\textbf{UniHR$_{Joint}$}    &\textbf{692.7}	&\textbf{0.489}	&0.409 &\textbf{0.530}	&\textbf{0.629}	&\textbf{686.5}	&\textbf{0.493}	&\textbf{0.418}	&\textbf{0.536}	&\textbf{0.632}	&\textbf{493.6}	&\textbf{0.317}	&\textbf{0.224}	&\textbf{0.348}	&0.503   \\ \bottomrule
\end{tabular}}
\vspace{-1.5mm}
\caption{Results of separate training and joint training on hybrid KG dataset \textbf{\textit{wikimix}}, where identical entities and relations share the same embeddings. WikiPeople$^{*}$ and wikidata12k$^{*}$ represent the filtered test sets.}\label{hybrid_form}
\vspace{-3mm}
\end{table*}
\begin{table}
\centering
\renewcommand{\arraystretch}{1.0} 
\resizebox{0.48\textwidth}{!}{
\begin{tabular}{lcccc|cccc}
\toprule
     & \multicolumn{2}{c}{\textbf{FBH}}                     & \multicolumn{2}{c}{\textbf{DBHE}} & \multicolumn{2}{c}{\textbf{FBH}}          & \multicolumn{2}{c}{\textbf{DBHE}}                         \\ \cmidrule(lr){2-5}\cmidrule(lr){6-9}

\multicolumn{1}{l}{\textbf{Model}}                      & MRR            & H@10                      & MRR            & \multicolumn{1}{c}{H@10}              & MRR            & H@10              & MRR            & H@10      \\ \cmidrule(lr){2-9} 
\multirow{1}{*}{}     & \multicolumn{4}{c|}{\textit{Base link prediction}}  & \multicolumn{4}{c}{\textit{Triple prediction}} \\ \midrule
\textbf{UniHR}          & 0.401          & 0.619       & 0.296          & 0.448     & 0.946          & 0.993       & \textbf{0.862}          & 0.987\\
\textbf{UniHR}$_{Joint}$         & \textbf{0.404}          & \textbf{0.632}       & \textbf{0.298}          & \textbf{0.454}    & \textbf{0.949}          & \textbf{0.994}       & 0.860          & \textbf{0.989}\\ \bottomrule

\end{tabular}}
\vspace{-1.5mm}
\caption{Results of separate and joint training on NKG.}\label{hybrid_task}
\vspace{-3mm}
\end{table}

\section{B More Related Works}

\paragraph{Link Prediction on Triple-based KGs.}Most existing techniques in KG representation learning are proposed for triple-based KGs. Among these techniques, knowledge graph embedding (KGE) models \cite{transe,rotate} have received extensive attention due to their effectiveness and simplicity. The idea is to project entities and relations in the KG to  low-dimensional vector spaces, utilizing KGE scoring functions to measure the plausibility of triples in the embedding space. Typical methods include TransE \cite{transe}, RotatE \cite{rotate}, and ConvE \cite{conve}.

Depending on the KGE model alone has limitation of capturing complex graph structures, whereas augmenting global structural information with a graph neural network (GNN) \cite{CompGCN,ma-gnn,zhanghe} proves to be an effective approach for enhancement. The paradigm of combining GNN as encoder with KGE scoring function as decoder helps to enhance the performance of KGE scoring function. These GNN methods design elaborate message passing mechanisms to capture the global structural features. 
Typically, CompGCN \cite{CompGCN} aggregates the joint embedding of entities and relations in the neighborhood via a parameter-efficient way and MA-GNN \cite{ma-gnn} learns global-local structural information based on multi-attention. These methods achieve impressive results on triple-based KGs but are challenging to generalize to beyond-triple KGs. Additionally, these GNNs usually focus more on modeling global information while neglecting local information.

\paragraph{Why We Need Unified Representation?}Firstly, real-world data sources naturally contain heterogeneous data forms~\cite{ska-bench}. For example, Wikidata~\cite{wikidata} includes triple-based facts, hyper-relational facts, temporal facts, and even nested facts, financial reports and research papers usually contain various modalities such as tables, images, and text. Unified representation aims to integrate multiple heterogeneous data within a single semantic structure, avoiding the need for cross-type data conversion or independent processing during use. In many real-world scenarios (e.g., question answering, recommendation, or transfer learning), the research community is constantly exploring the benefits of unified representations. For question answering (QA)~\cite{rtqa,eog}, TrustUQA~\cite{trustuqa} demonstrates a unified representation allows the knowledge base to contain a broader range of knowledge types, thereby simplifying the burden of the knowledge retrieval module and enabling the QA system~\cite{prgb} to cover a wider variety of questions and answers. For recommendation systems, DURation~\cite{unified_rec} with unified representation effectively eliminate information redundancy caused by separate module designs and help capture cross-dimensional relational knowledge, thus improving the comprehensiveness and accuracy of recommendations. Moreover, UniHR not only meets the requirements of real-world KG reasoning tasks (involving multiple types of facts), but we also believe that unified representation itself is beneficial for transfer learning. It not only facilitates transfer learning across different types of KGs, but also enables pre-training across multiple KG types (as demonstrated in Section 5.4). This further promotes more effective pre-training and transfer learning in the field of knowledge graphs.


\section{C Detailed Results of Our ``Potential''}
Table~\ref{hybrid_form} shows the results of ``Joint Learning on Different Tasks of KGs'' section. Table~\ref{hybrid_task} shows the results of ``Joint Learning on Different Types of KGs'' section.

\section{D Dataset Details}
Table \ref{tabc1} shows the details of the two hyper-relational knowledge graph  (HKG) benchmark datasets: WikiPeople, WD50K three nested factual knowledge graph (NKG) benchmark datasets: FBH, FBHE, DBHE, and the temporal knowledge graph (TKG) benchmark dataset wikidata12k. 

Among them, WikiPeople is a dataset derived from Wikidata \cite{wikidata} concerning entities type “human”. WikiPeople filter out the elements which have at least 30 mentions as key-value pairs. WD50K is a high-quality dataset extracting from Wikidata statements and avoiding the potential data leakage which allows triple-based models to memorize main fact in the H-Facts of test set. The ``with Q(\%)" column in Table \ref{tabc1} denote the number of facts with auxiliary key-value pairs and the ``Arity" column denote range of the number of entities in H-Facts. 

Nested factual knowledge graph datasets FBH and FBHE \cite{bive} are constructed based on FB15k237 from Freebase \cite{bollacker2008freebase} while DBHE is based on DB15K from DBpedia \cite{DBpedia}. FBH contains nested facts that can be only inferred inside the atomic facts, while FBHE and DBHE contain externally-sourced nested relation crawling from Wikipedia articles, e.g., NextAlmaMater and SucceededBy.

Temporal knowledge graph dataset wikidata12K is also a subset of Wikidata \cite{wikidata}, which represents the time information $\tau\in \mathcal{T}$ as time intervals.

\section{E Hyperparameter Settings}\label{app:hyper}
Here, we show the hyperparameter details for each link prediction task. To be specific, we tune the learning rate using the range $\left\{ 0.0001,0.0005,0.001 \right\} $, the embedding dim using the range $\left\{ 50,100,200,400 \right\} $, the GNN layer using the range $\left\{ 1,2,3 \right\}$ and dropout using the range $\left\{ 0.1,0.2,0.3,0.4 \right\}$. Additionally, we use smoothing label in the training phase from range $\left\{ 0.1,0.2,0.3\right\}$. The best hyperparameters obtained from the experiments are presented in Table \ref{tabd1}.

\end{document}